\documentclass{article}

\PassOptionsToPackage{numbers,compress}{natbib}

\usepackage[preprint]{neurips_2026}
\usepackage{multirow}

\usepackage[utf8]{inputenc} 
\usepackage[T1]{fontenc}    
\usepackage{hyperref}       
\usepackage{url}            
\usepackage{array}          
\usepackage{booktabs}       
\usepackage{amsmath}        
\usepackage{amsfonts}       
\usepackage{nicefrac}       
\usepackage{microtype}      
\usepackage[table]{xcolor}  

\usepackage{graphicx}       
\usepackage{caption}        
\usepackage{wrapfig}        
\usepackage{enumitem}
\usepackage{tikz}
\usetikzlibrary{positioning,arrows.meta,calc,fit,decorations.pathreplacing}
\usepackage{pifont}

\usepackage[most,skins,theorems]{tcolorbox}
\tcbset{
  takeawaysbox/.style={
    title=Takeaway, 
    colback=teal!5!white,
    colframe=teal!65!black,
    fonttitle=\bfseries\sffamily\scshape\small,
    coltitle=white,
    colbacktitle=teal!50!black,
    enhanced,
    attach boxed title to top left={xshift=2mm, yshifttext=-1mm, yshift=-2mm},
    boxed title style={
        outer arc=2mm,
        arc=1.5mm, 
        colframe=teal!65!black,
        colback=teal!50!black,
    },
    width=\linewidth,
    arc=3mm,
    boxrule=0.8pt, 
  }
}

\newcommand{\trice}{\textsc{Trice}}

\NewDocumentCommand{\yuchen}
{ mO{} }{\textcolor{blue}{\textsuperscript{\textit{yuchen}}\textsf{\textbf{\small[#1]}}}}

\NewDocumentCommand{\yun}
{ mO{} }{\textcolor{green}{\textsuperscript{\textit{yun}}\textsf{\textbf{\small[#1]}}}}

\title{Teaching Thinking Models to Reason with Tools:\\A Full-Pipeline Recipe for Tool-Integrated Reasoning}

\author{%
  Qianjia Cheng$^{1,2*}$, Yuchen Zhang$^{2,3*}$,  Zhilin Wang$^{2,6}$,  Yuxin Zuo$^{5}$,  Shunkai Zhang$^{2,3}$, \\ \textbf{Yuchen Fan}$^{7}$,  \textbf{Yu Qiao}$^{2}$, \textbf{Bowen Zhou}$^{2,5}$, \textbf{Ning Ding}$^{5}$$^\dagger$, \textbf{Yu Cheng}$^{2,4}$$^\dagger$, \textbf{Yun Luo}$^{2}$$^\dagger$, \textbf{Ganqu Cui}$^{2}$$^\dagger$ \\
  $^1$ Zhejiang University \quad $^2$ Shanghai AI Laboratory \quad $^3$ Peking University \quad  \\ $^4$ The Chinese University of Hong Kong \quad $^5$ Tsinghua University \\ $^6$ University of Science and Technology of China \quad $^7$ Shanghai Jiao Tong University \\
  \texttt{chengqianjia@zju.edu.cn} \\
}

\begin{document}

\maketitle
\begingroup
\renewcommand{\thefootnote}{}
\footnotetext{$^*$ Equal contribution. ~~~$^\dagger$ Corresponding authors. ~~~Our models and dataset are coming soon.}
\endgroup
\begin{abstract}
Tool-integrated reasoning (TIR) offers a direct way to extend thinking models beyond the limits of text-only reasoning. Paradoxically, we observe that tool-enabled evaluation can degrade reasoning performance even when the strong thinking models make almost no actual tool calls. In this paper, we investigate how to inject natural tool-use behavior into a strong thinking model without sacrificing its no-tool reasoning ability, and present a comprehensive TIR recipe. We highlight that (i) the effectiveness of TIR supervised fine-tuning (SFT) hinges on the learnability of teacher trajectories, which should prioritize problems inherently suited for tool-augmented solutions; (ii) controlling the proportion of tool-use trajectories could mitigate the catastrophic forgetting of text-only reasoning capacity; (iii) optimizing for \textit{pass@k} and response length instead of training loss could maximize TIR SFT gains while preserving headroom for reinforcement learning (RL) exploration; (iv) a stable RL with verifiable rewards (RLVR) stage, built upon suitable SFT initialization and explicit safeguards against mode collapse, provides a simple yet remarkably effective solution. When applied to Qwen3 thinking models at 4B and 30B scales, our recipe yields models that achieve state-of-the-art performance
in a wide range of benchmarks among open-source models, such as 96.7\% and 99.2\% on AIME 2025 for 4B and 30B, respectively.
\end{abstract}

\section{Introduction}
\label{sec:intro}

\begin{figure}[h]
    \centering
    \includegraphics[width=\linewidth]{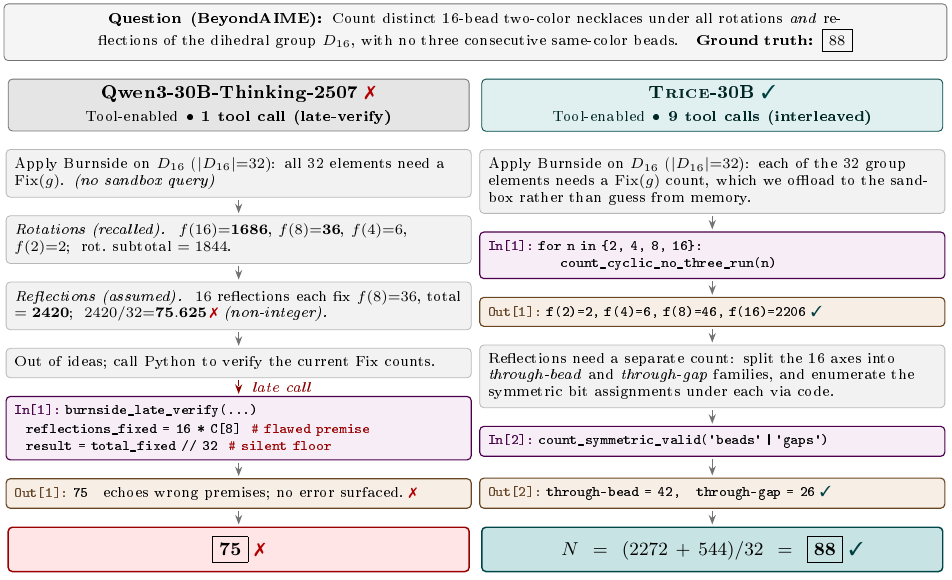}
    \caption{%
    \textbf{The same problem, two policies for invoking the tool.}
    Grey boxes denote text-only reasoning; \colorbox[RGB]{246,237,246}{\texttt{In[k]}} cells indicate tool calls, and \colorbox[RGB]{247,238,230}{\texttt{Out[k]}} cells represent tool responses.
    \textbf{Left:} Qwen3-30B-Thinking-2507 treats the Python sandbox as a final-pass verifier. After a text-only Burnside derivation yields the inconsistent value $2420/32 = 75.625$, it makes a single late-stage call that hard-codes the flawed assumptions (e.g.\ \texttt{reflections\_fixed = 16*C[8]}). A silent integer floor masks the error, leading to the incorrect result $\boxed{75}$\,\textcolor{red!70!black}{\ding{55}}.
    \textbf{Right:} \trice-30B interleaves textual reasoning with code execution, feeding each intermediate result back into the same Burnside framework, and correctly obtains $\boxed{88}$\,\textcolor{teal!50!black}{\ding{51}}.\;
    }
    \label{fig:intro_case}
    \vspace{-10pt}
\end{figure}

Current thinking models have demonstrated that scaling the reasoning
depth can substantially improve performance on difficult tasks with pure natural language~\citep{openai2024gpt4technicalreport, deepseekai2025deepseekr1incentivizingreasoningcapability, cui2025process, zhang2025survey, zeng2025simplerlzooinvestigatingtamingzero}.
However, in more complex scenarios that frequently require formal reasoning (e.g., numerical computation, symbolic manipulation, and systematic case enumeration), text-only reasoning often struggles, making it both costly and less reliable~\citep{li2026discovery, zhang2025tool, cort}.
TIR provides a natural way out. By allowing the model to interleave natural-language deliberation with code execution~\citep{pal, retool}, TIR combines the flexibility of text-only reasoning with the precision and expertise of the external feedback.

The potential of TIR is clear in principle, yet simply providing a thinking model with access to a code executor does not guarantee interleaved reasoning. As shown in Table~\ref{tab:motivation}, across 400 tool-enabled rollouts, Qwen3 thinking models make almost no tool calls, continuing with text-only reasoning. For example, as illustrated in Figure~\ref{fig:intro_case}, the model invokes the executor only once at the very end to verify the final answer, rather than leveraging computation to guide intermediate reasoning steps. This behavior reflects the delayed-code pattern \citep{cort, du2025generalizable}, where code is applied too late or even not at all, rather than being integrated into the reasoning loop. We identify the core bottleneck as the lack of a reliable policy for determining when reasoning should hand off to computation and how to incorporate the results back into the reasoning process.

\begin{table}[t]
  \caption{Naively enabling a Python tool does not induce interleaved thinking
  on BeyondAIME. Calls$_t$ and Len$_t$ are measured
  only on TIR trajectories.}
  \vspace{1mm}
  \label{tab:motivation}
  \centering
  \small
  \begin{tabular}{lcccc}
    \toprule
    Model & Acc$_{n/t}$ & Pass@4$_{n/t}$ & Calls$_t$ & Len$_t$ \\
    \midrule
    Qwen3-4B-Thinking & 54.3 / 50.3 & 66 / 62 & 0.07 & 15.3k \\
    Qwen3-30B-Thinking & 66.0 / 58.0 & 79 / 68 & 0.00 & 17.1k \\
    Qwen3-30B-Instruct & 51.3 / 39.5 & 66 / 58 & 1.38 & 9.7k \\
    \bottomrule
  \end{tabular}
  \vspace{-10pt}
\end{table}

To address this challenge, prior work has explored several approaches, including fine-tuning on tool-using trajectories and reinforcement learning that rewards successful tool invocation \citep{pal,retool,cort,simpletir}. 
However, these approaches often overlook the structural fragility of reasoning chains during tool integration. Our empirical analysis in Table~\ref{tab:motivation} reveals a critical bottleneck: unlike standard instruct models, thinking models exhibit a sharp performance drop when prompted with tools, indicating that their internal reasoning logic is easily disrupted. It indicates that teaching a strong thinking model to think with tools without sacrificing its text-only reasoning performance requires more effort. To resolve this inherent instability, we move beyond scattered techniques and propose a systematic, full-pipeline recipe spanning data preparation, SFT, the transition from SFT to RL, and RL itself. The recipe is
organized as the following components:

\begin{itemize}[leftmargin=*, itemsep=0.5em, topsep=0.5em]
\item \textbf{Data Engineering:} For each training stage, we carefully build data collection and filtering pipelines. 
For teacher prompts, we find that \textit{tool-advantaged} problems could better encourage tool use.
For teacher trajectories, we find that 1) pattern learnability matters a lot, and 2) mixing TIR and text-only reasoning trajectories helps prevent catastrophic degradation of text-only reasoning. 
Based on these findings, we construct high-quality teacher trajectories for SFT training. Moreover, we construct a challenging RLVR dataset to naturally elicit tool usage patterns in RL training.
    
\item \textbf{Stage Coordination:} We identify that what the student learns during SFT evolves through form, substance, and noise, and argue that SFT training should prioritize downstream RL potential over standard supervised loss. To this end, we utilize \textit{pass@k} performance and rollout length to determine the optimal SFT extent for maximizing subsequent RL gains.
    
\item \textbf{Reinforcement Learning:} We show that a suitable SFT initialization and on-policy training stabilize multi-turn tool use, improving both with-tool and no-tool performance. We forgo the format rewards and leverage outcome-based rewards only with challenging questions to naturally elicit tool use. Applied to Qwen3 thinking models at 4B and 30B scales, the resulting models, capable of \textbf{T}extual \textbf{R}easoning \textbf{I}nterleaved with \textbf{C}ode \textbf{E}xecution, termed as \trice-4B and \trice-30B, achieve state-of-the-art performance among open-source models of this scale in a wide range of benchmarks, such as reaching 96.7\% and 99.2\% on AIME 2025, respectively.
\end{itemize}
\section{Related Work}
\label{sec:related}

\paragraph{Reinforcement learning for large reasoning models.}

Reinforcement learning has made remarkable progress for building large reasoning models~\citep{openai2024gpt4technicalreport, cui2025process, cui2025entropy, yuan2025f, zuo2025ttrl}. By optimizing with verifiable rewards, it encourages LLMs to scale up their reasoning processes, thereby advancing the frontier of intelligence~\citep{chen2025p1, deepseekai2025deepseekr1incentivizingreasoningcapability, zhang2025survey, luo2026p1, he2025justrl}. However, text-only reasoning is inherently limited for problems that require exact numerical computation or rigorous verification. Therefore, external tools that compensate for these limitations can be naturally integrated into the reasoning process~\citep{yu2025demystifying, du2025nemotron, dong2025agentic, understandingtir, wang2025information}.

\paragraph{Tool-integrated reasoning.}
Although introducing executable tools into reasoning has shown clear
feasibility, injecting this ability into an already strong thinking model is not
straightforward: it requires reshaping a largely fixed policy~\citep{li2026discovery,zhang2026aster, xia2025agent0}. Prior work
focuses on different parts of this problem, including the cold-start tool
trajectories, code placement, theoretical support expansion, RL stability, long-CoT tool discovery, interaction-dense priors, and tool-use pattern
selection \citep{retool,cort,understandingtir,simpletir, shang2025rstar2}.
Beyond the above work, we aim to convert an already strong thinking model into a reliable
TIR model while maintaining or improving its no-tool performance through a full-pipeline recipe.

\section{Preliminary}
\label{sec:prelim}
\subsection{Reinforcement Learning with Verifiable Reward}
Given a problem
$x \sim \mathcal{D}$, an LLM policy $\pi_\theta$ autoregressively generates a
serialized trajectory $y=(y_1,\ldots,y_T)$ and receives a terminal reward
$r(x,y)$ from a verifier. The standard RL objective is
\begin{equation}
    J(\theta)=
    \mathbb{E}_{x \sim \mathcal{D},\, y \sim \pi_\theta(\cdot \mid x)}
    \left[r(x,y)\right].
\end{equation}
Policy-gradient methods estimate its gradient by assigning an advantage to
generated tokens \citep{sutton2018reinforcement}:
\begin{equation}
    \nabla_\theta J(\theta)=
    \mathbb{E}\left[
    \sum_{t=1}^{T}
    \nabla_\theta \log \pi_\theta(y_t \mid x,y_{<t}) A_t
    \right].
\end{equation}
With only a trajectory-level reward, REINFORCE can use $A_t=r(x,y)$ for all
tokens. In practice, group-normalized variants such as GRPO sample $K$
responses $\{y^{(k)}\}_{k=1}^{K}$ for the same prompt and use a normalized
trajectory advantage,
\begin{equation}
    A_t^{(k)} =
    \frac{r(x,y^{(k)})-\operatorname{mean}_{j} r(x,y^{(j)})}
    {\operatorname{std}_{j} r(x,y^{(j)})+\epsilon},
\end{equation}
which reduces variance and supports learning from sparse verifiable rewards
\citep{deepseekmath}.

\subsection{Tool-Integrated Reasoning}
Tool-integrated reasoning uses the same objective, but the trajectory $y$ is
produced through an interactive process before being verified. At interaction
turn $\tau$, the state is $s_\tau=(x,H_\tau,E_\tau)$, where $H_\tau$ is the
current interaction history and $E_\tau$ is the sandbox state. The policy takes
an action $a_\tau$ that is either natural-language reasoning or executable
code. Text actions are appended directly to the history. Code actions are
executed by the sandbox, and it returns an observation $o_\tau$, such as printed
output or an error message, so the next state is
\begin{equation}
    s_{\tau+1} =
    (x, H_\tau \oplus a_\tau \oplus o_\tau, E_{\tau+1}).
\end{equation}
When the model emits a final answer or reaches the turn budget, the final
history is serialized as $y$ and scored by the same verifier reward $r(x,y)$.
We use a stateful sandbox, so variables, imports, and intermediate
computations persist across tool calls; this lets the model inspect, revise,
and extend prior computations rather than restart from an empty environment.
The no-tool setting is the special case in which code actions are disallowed,
so $y$ is generated entirely through language.
\section{Data Engineering for TIR}
\label{sec:data-engineering}

In this section, we detail the SFT data curation pipeline of our TIR recipe, focusing on the strategic selection of teacher models and the SFT trajectories.


\vspace{-10pt}

\subsection{Teacher Model Selection}
\label{subsec:teacher-selection}

\begin{tcolorbox}[takeawaysbox, title=Takeaway 1]
For TIR SFT, teacher selection should account for the learnability of tool-use patterns, not teacher accuracy alone.
\end{tcolorbox}

When selecting a teacher model for TIR SFT, we argue that downstream task performance alone is insufficient~\citep{guha2025openthoughts}, since the teacher’s tool-use pattern directly shapes the reasoning strategy learned by the student. Considering the cost and tool-use capability, we mainly compare GPT-OSS-120B~\citep{openai2025gptoss120bgptoss20bmodel} and MiniMax-M2.7~\citep{minimax2026m} from the frontier models~\citep{kimiteam2026kimik25visualagentic, deepseekai2025deepseekv32pushingfrontieropen, openai2025gptoss120bgptoss20bmodel, minimax2026m}.

\begin{table}[t]
    \caption{Teacher comparison. Calls represent tool calls per trajectory. Len means response length. Snippet is code-block length in tokens. Tool Use is the fraction of problems with ${\geq}1$ tool calls. }
      \vspace{1mm}
    \label{tab:teacher-compare}
    \centering\small
    \setlength{\tabcolsep}{5pt}
    \begin{tabular}{l ccc c cccc}
      \toprule
      & \multicolumn{3}{c}{Accuracy (\%)} & & \multicolumn{4}{c}{Trajectory Statistics} \\
      \cmidrule(lr){2-4} \cmidrule(lr){6-9}
      Model & AIME25 & HMMT25 & IMOAnswerBench && Calls & Len & Snippet & Tool Use \\
      \midrule
      GPT-OSS-120B  & 99.2 & 92.5  & 77.3 && 12.2 & 21.6K & 137  & 99.7\% \\
      MiniMax-M2.7  & 99.2 & 91.7 & 76.3 && \phantom{0}5.1 & 27.4K & 325 & 66.4\% \\
      \bottomrule
    \end{tabular}
    \vspace{-12pt}
\end{table}

As shown in Table~\ref{tab:teacher-compare}, the two models achieve comparable performance, but exhibit markedly different tool-use styles despite being equipped with the same stateful sandbox environment described in Section~\ref{sec:prelim}. Specifically, GPT-OSS-120B follows a \emph{high-frequency, lightweight} pattern: it generates short code snippets and incrementally builds on variables, functions, and intermediate results from previous tool calls, thereby effectively exploiting the persistent sandbox state. In contrast, MiniMax-M2.7 follows a more typical stateless \emph{low-frequency, heavyweight} pattern: it tends to generate longer self-contained programs with repeated definitions, and shows a substantially lower tool-use rate.

\begin{wraptable}{r}{0.40\textwidth}
  \vspace{-1.2em}
  \caption{The performance (\%) of 4B student models trained with each teacher's trajectories on BeyondAIME.}
  \label{tab:teacher-student}
  \centering\small
  \setlength{\tabcolsep}{4pt}
  \begin{tabular}{l ccc}
    \toprule
    Teacher & Acc & Calls & Len \\
    \midrule
    GPT-OSS-120B & 55.8 & 29.3 & 37.2K \\
    MiniMax-M2.7 & 17.8 & 1.2 & 68.3K \\
    \bottomrule
  \end{tabular}
\end{wraptable}

To assess whether these tool-use patterns are learnable by the student model, we fine-tune Qwen3-4B-Thinking-2507 under the same setup using 16K trajectories from each teacher. Table~\ref{tab:teacher-student} shows that with the \emph{high-frequency, lightweight} teacher, the student yields higher accuracy and shorter responses. We attribute this to three factors. First, lightweight snippets decompose computation into simple, single-purpose steps that a 4B model can generate more reliably, whereas the \emph{low-frequency, heavyweight} style requires longer self-contained programs that are more vulnerable to autoregressive error accumulation~\citep{jiang2025rocode}, often leading to repeated debugging and budget exhaustion. Second, \emph{high-frequency, lightweight} calls better support interleaved reasoning: the model can offload small-case exploration whenever needed, and immediately incorporate execution feedback into subsequent reasoning. In contrast, heavyweight calls require substantial textual pre-planning, making the underlying TIR strategy harder for the student model to master. Third, lightweight snippets naturally reuse variables, functions, and execution states in the stateful sandbox, improving both token and execution efficiency. We therefore select GPT-OSS-120B as our teacher model.

\subsection{SFT Data Selection}
\label{subsec:data-composition}
\begin{tcolorbox}[takeawaysbox, title=Takeaway 2]
Selecting problems amenable to tool-augmented solutions (termed as \textit{tool-advantaged} problems) as teacher inputs can better elicit students to use tools from teacher trajectories.
\end{tcolorbox}

After selecting the teacher, the next question is how to choose the prompts and trajectories. We answer this question from the following perspectives.

\textbf{Prompt selection.} First, we select the prompts for the teacher model to produce trajectories more suitable for students. The quality of TIR supervision varies across problems. Problems whose difficulty lies in computational structure tend to benefit more from tool access, while those centered on conceptual insight benefit less. Moreover, the teacher's tool-use policy itself is imperfect and may not leverage tools effectively on every problem. To select prompts likely to yield more informative trajectories, we compare the teacher's average accuracy over $k$ rollouts with and without tool access per problem and retain those where TIR accuracy substantially exceeds the text-only reasoning, termed \textit{tool-advantaged} problems. As Table~\ref{tab:prompt-selection} shows, the model trained on trajectories from tool-advantaged prompts performs much better with a 5.4\% advantage.

\begin{table}[t]
  \centering
  \begin{minipage}[t]{0.48\textwidth}
    \centering\small
    \caption{With-tool accuracy (\%) of student models with and without prompt selection.}
      \vspace{1mm}
    \label{tab:prompt-selection}
    \setlength{\tabcolsep}{5pt}
    \begin{tabular}{l cc}
      \toprule
      Prompts & HMMT25 & BeyondAIME \\
      \midrule
      Base (no SFT) & 58.3 & 50.3 \\
      Random & 69.6 & 52.3 \\
      Tool-adv. & \textbf{75.0} & \textbf{53.5} \\
      \bottomrule
    \end{tabular}
  \end{minipage}
    \hfill
    \begin{minipage}[t]{0.48\textwidth}
    \centering\small
    \caption{With-tool accuracy (\%) under different trajectory composition strategies.}
      \vspace{1mm}
    \label{tab:traj-composition}
    \setlength{\tabcolsep}{4pt}
    \begin{tabular}{l cc}
      \toprule
      Configuration & HMMT25 & BeyondAIME \\
      \midrule
      TIR traj., 32K & 75.0 & 53.5 \\
      Mixed traj., 32K & 78.3 & 53.0 \\
      Mixed traj., 16K & \textbf{82.5} & \textbf{53.3} \\
      \bottomrule
    \end{tabular}
  \end{minipage}
  \vspace{-5pt}
\end{table}

\begin{tcolorbox}[takeawaysbox, title=Takeaway 3]
To preserve the student model's native text-only reasoning capability, mix text-only trajectories into the TIR set to avoid catastrophic forgetting
\end{tcolorbox}

\begin{wraptable}{r}{0.43\textwidth}
  \vspace{-1.2em}
  \caption{Text-only BeyondAIME accuracy and truncation rate (\%) after SFT.}
  \label{tab:sft-truncation}
  \centering\small
  \setlength{\tabcolsep}{5pt}
  \begin{tabular}{l cc}
    \toprule
    Configuration & Accuracy & Trunc. Rate \\
      \midrule
      Base & 54.3 & 0 \\
      TIR traj., 32K & 34.1 & 44 \\
      Mixed traj., 32K & 49.0 & 19 \\
    \bottomrule
  \end{tabular}
\end{wraptable}

\textbf{Mix TIR trajectories with text-only reasoning ones.} Despite the strong TIR performance of TIR-only SFT, the same models exhibit severe text-only reasoning degradation: accuracy drops across all benchmarks, as shown in Table~\ref{tab:sft-truncation}. Moreover, the text-only truncation rate surges to ${\sim}$40\%. We study these text-only reasoning trajectories (e.g., Appendix~\ref{app:skill-antagonism}) and attribute this to a cross-mode negative transfer: TIR-only SFT causes the model to internalize a code-centric reasoning loop that persists even when no tool is available. The model attempts to simulate code execution in natural language, for example, writing programs, fabricating outputs, and iterating on hallucinated results---ultimately exhausting the token budget without mathematical progress~\citep{du2025nemotron, gao2025train}. To alleviate this phenomenon, we developed a simple but effective approach: mixing text-only trajectories into TIR trajectories. In this way, the text-only reasoning performance recovers (Table~\ref{tab:sft-truncation}) while TIR accuracy is still preserved (Table~\ref{tab:traj-composition}).

\begin{tcolorbox}[takeawaysbox, title=Takeaway 4]
Filtering out overlong teacher trajectories not only improves efficiency for downstream RL, but also prevents the student from falling into the trap of imitating reasoning length.
\end{tcolorbox}

\textbf{Overlong filtering.} Considering downstream RL rollout efficiency, we restrict trajectories to 16K tokens to control the rollout length of the student models. Surprisingly, it improves HMMT25 accuracy from 78.3\% to 82.5\%, while maintaining comparable BeyondAIME performance as presented in Table~\ref{tab:traj-composition}. It implies that filtering out overly long trajectories during SFT helps prevent the student model from exploiting length as a spurious signal, encouraging it to focus more on learning the substance of TIR.

\section{Stage Coordination: From SFT to RL}
\label{sec:training-dynamics}

Based on the curated SFT data, this section details a principled execution of the fine-tuning process to fully unlock its potential within the holistic training pipeline.

\begin{tcolorbox}[takeawaysbox, title=Takeaway 5]
During TIR SFT, what the student learns evolves across form, substance, and noise. We determine the training extent based on \textit{pass@k} performance and rollout length.
\end{tcolorbox}

\subsection{TIR SFT Dynamics}
\label{subsec:sft-dynamics}

\begin{figure}[t]
  \centering
  \begin{minipage}[t]{0.245\linewidth}
    \centering
    \includegraphics[width=\linewidth]{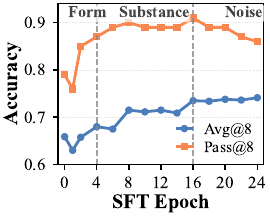}
    \vspace{-0.3em}
    \textbf{(a)} Accuracy
  \end{minipage}\hfill
  \begin{minipage}[t]{0.245\linewidth}
    \centering
    \includegraphics[width=\linewidth]{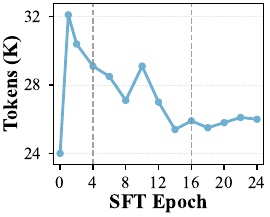}
    \vspace{-0.3em}
    \textbf{(b)} Response length
  \end{minipage}\hfill
  \begin{minipage}[t]{0.245\linewidth}
    \centering
    \includegraphics[width=\linewidth]{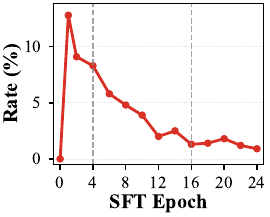}
    \vspace{-0.3em}
    \textbf{(c)} Length trunc. rate
  \end{minipage}\hfill
  \begin{minipage}[t]{0.245\linewidth}
    \centering
    \includegraphics[width=\linewidth]{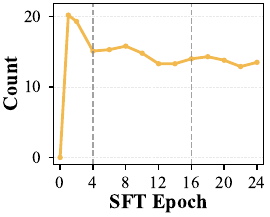}
    \vspace{-0.3em}
    \textbf{(d)} Tool call counts
  \end{minipage}
  \caption{\textbf{TIR SFT dynamics of the 30B model.} We perform SFT on Qwen3-30B-A3B-Thinking-2507 using expert data generated by GPT-OSS-120B. The observed learning curve (measured on BeyondAIME) demonstrates a "form--substance--noise" progression.}
  \label{fig:sft-tir-dynamics}
  \vspace{-5pt}
\end{figure}

\begin{wrapfigure}{r}{0.3\linewidth}
\vspace{-1.2em}
\centering
\includegraphics[width=\linewidth]{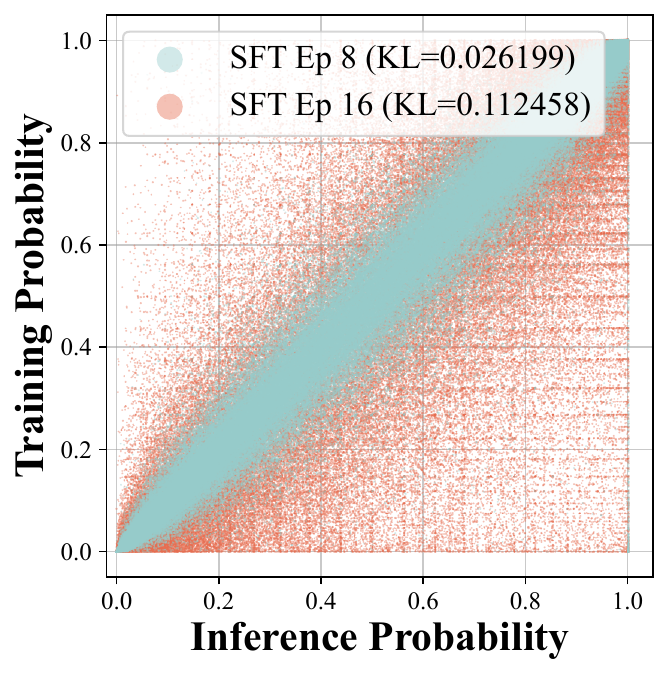}
\caption{The token probabilities assigned by the inference and training engines to the same rollouts.}
\label{fig:late-sft-mismatch}
\vspace{-0.7em}
\end{wrapfigure}

During TIR SFT, the student model follows a form–substance–noise learning progression, which we characterize as \textit{stage 1-3}. In the early stage, it quickly acquires the format of tool invocation, causing tool-use frequency to rise sharply, which yet does not yield effective TIR. As shown in Figure~\ref{fig:sft-tir-dynamics}, performance drops while the truncation rate increases. Case in Appendix~\ref{app:early-sft-form} further shows that early tool calls are often inefficient symbolic exploration, repeated verification, or loops, which accumulate errors and produce overly long reasoning trajectories. It suggests that the model is primarily imitating the form of TIR in this stage (\textit{stage 1}). As training proceeds, tool-use frequency stabilizes, and tool calls more often provide useful intermediate evidence. Correspondingly, both pass@1 and pass@8 improve, indicating that the model moves beyond exploiting the superficial signal of tool-call frequency and begins to learn the substance of TIR, adapting tool use to its reasoning prior. This resembles the degradation–recovery dynamics observed in long-CoT SFT~\citep{zhang2026onpolicyrlmeetsoffpolicy, luo2025through}(\textit{stage 2}). Finally, after TIR behavior has been sufficiently internalized and useful supervision diminishes, the following SFT gradually overfits the teacher-side noise (e.g., rollout length), leading the performance to saturate or even decline (\textit{stage 3}).

\subsection{Identifying RL-Ready SFT Checkpoints}
\label{subsec:checkpoint-selection}

\begin{figure}[t]
  \centering
  \begin{minipage}[t]{0.32\linewidth}
    \centering
    \includegraphics[width=\linewidth]{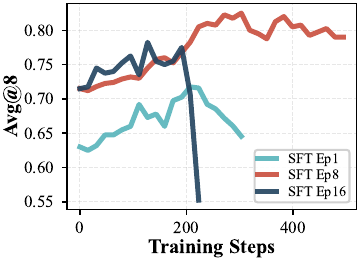}
    \vspace{-0.3em}
    \textbf{(a)} Accuracy
  \end{minipage}\hfill
  \begin{minipage}[t]{0.32\linewidth}
    \centering
    \includegraphics[width=\linewidth]{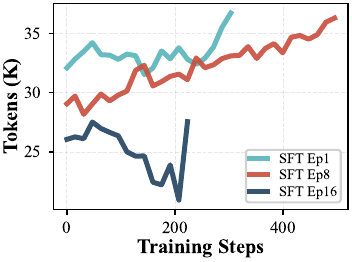}
    \vspace{-0.3em}
    \textbf{(b)} Response length
  \end{minipage}\hfill
  \begin{minipage}[t]{0.32\linewidth}
    \centering
    \includegraphics[width=\linewidth]{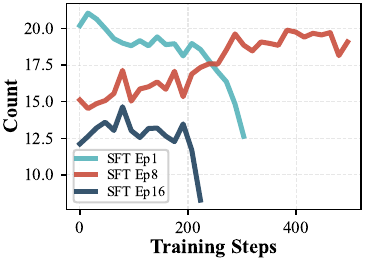}
    \vspace{-0.3em}
    \textbf{(c)} Tool call counts
  \end{minipage}
    \vspace{1mm}
  \caption{The RL training dynamics of the 30B model, EP is short for Epoch. Ep 1 is the model in \textit{stage 1}, Ep 8 is the model in \textit{stage 2}, while Ep 16 is the model in \textit{stage 3}.}
  \label{fig:rl-from-sft-checkpoints}
  \vspace{-10pt}
\end{figure}

The learning pattern observed in the student model informs our choice of the SFT endpoint. Specifically, we aim to select a checkpoint that has sufficiently learned the substance of TIR, but has not yet absorbed excessive noise, as the initialization for RL. As discussed in Section~\ref{subsec:sft-dynamics}, we find that the transitions between stages are closely associated with two diagnostics: \textit{pass@k} performance and rollout length. We therefore use these metrics to guide checkpoint selection.

To validate this criterion, we initialize RL from checkpoints at different stages, namely epochs 1, 8, and 16. As shown in Figure~\ref{fig:rl-from-sft-checkpoints}, checkpoints from both \textit{stage 1} and \textit{stage 3} eventually collapse during RL training. The failure of \textit{stage 1} is straightforward: the model has not yet learned an effective TIR strategy, but already invokes tools frequently. Under outcome-only rewards, RL further amplifies these erroneous patterns, ultimately leading to training collapse~\citep{li2025taming}.

The failure of \textit{stage 3} is more subtle, since its \textit{pass@1} performance is even higher than that of the stage-2 checkpoint. We therefore conduct a deeper analysis of the stage-3 model. As shown in Figure~\ref{fig:late-sft-mismatch}, a salient difference in RL is that, compared with the stage-2 model, the stage-3 model has a $\sim$5 times larger train–inference mismatch. We attribute this to partial overfitting to the teacher model in \textit{stage 3}, where the student begins to absorb teacher-side noise. Moreover, due to the off-policy nature of tool-return tokens in the context, the model becomes more prone to emitting extremely low-probability tokens, making RL optimization substantially more difficult~\citep{liu2025speed,zheng2025stabilizing}.

\subsection{Stable RL Training for TIR}
\label{subsec:rl-recipe}

\begin{tcolorbox}[takeawaysbox, title=Takeaway 6]
For the off-policy nature of tool-call returns, stability must be a central consideration in TIR RL. Using on-policy rollouts together with rollout routing replay is simple but necessary.
\end{tcolorbox}

In the RL stage, we do not introduce any reward that directly incentivizes tool invocation, since tuning its weight is difficult and may also induce reward hacking. Instead, we carefully construct a high-difficulty, IMO-level dataset from scratch for training, which is available at Appendix~\ref{app:rl-data}. 
Intuitively, tasks that exceed the inherent limits of text-only reasoning provide a stronger functional incentive for the model to invoke external tools.

\begin{figure}[t]
  \centering
  \begin{minipage}[t]{0.32\linewidth}
    \centering
    \includegraphics[width=\linewidth]{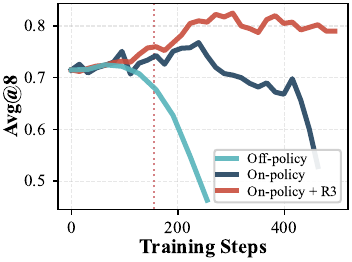}
    \vspace{-0.35em}
    \textbf{(a)} Acc on BeyondAIME
  \end{minipage}\hfill
  \begin{minipage}[t]{0.32\linewidth}
    \centering
    \includegraphics[width=\linewidth]{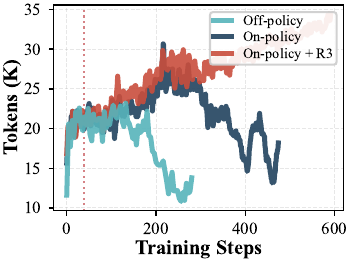}
    \vspace{-0.35em}
    \textbf{(b)} Training resp. length
  \end{minipage}\hfill
  \begin{minipage}[t]{0.32\linewidth}
    \centering
    \includegraphics[width=\linewidth]{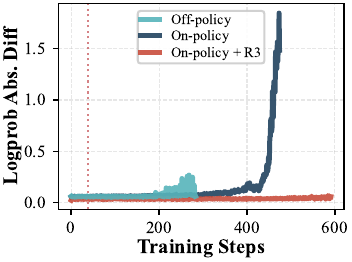}
    \vspace{-0.35em}
    \textbf{(c)} Train--rollout mismatch
  \end{minipage}
  \vspace{1mm}
  \caption{Comparison of different RL training strategies. An off-policy setting (four updates per rollout) leads to collapse. On-policy updates stabilize training, and adding rollout routing replay (R3) further improves the final score while keeping the train--inference mismatch controlled.}
  \label{fig:rl-stability-ablation}
  \vspace{-10pt}
\end{figure}

However, even under the simplest RLVR setup, training quickly collapses with off-policy updates, where each rollout batch is partitioned into four mini-batches for gradient updates. Surprisingly, simply switching to fully on-policy training could mitigate this issue, as shown in Figure~\ref{fig:rl-stability-ablation}. Moreover, adding rollout routing replay~\citep{ma2025stabilizing}, which reuses inference-time routing distributions during training, further improves stability and performance (Figure~\ref{fig:rl-stability-ablation}). These results highlight that, in the TIR setting, the off-policy nature of external tool returns makes it especially important to avoid additional distributional shifts and to prioritize RL training stability.

\section{Experiments}
\label{sec:results}

\subsection{Setup}
\label{sec:eval-protocol}

In the experiment, we compare against a broad set of open-source reasoning models. Unless otherwise noted, all models are evaluated under our unified protocol on five competition-level benchmarks: AIME\,2025~\citep{li2024numinamath}, HMMT\,2025~\citep{dekoninck2026matharena}, BeyondAIME~\citep{bytedance_beyondaime}, IMOAnswerBench~\citep{luong-etal-2025-towards}, and APEX\,2025~\citep{dekoninck2026matharena}. Every question is repeated 8 times to ensure reproducibility. We use a consistent configuration of 80K maximum rollout length and up to 128 tool calls in a stateful sandbox. 

\subsection{Main Results}

\begin{table*}[t]
    \caption{
        Performance comparison on competition-level mathematical benchmarks.
        Results are accuracy (\%) under each model's indicated inference setting.
        Models marked with $^\dagger$ are reported by concurrent TIR systems under their original protocols; all other results follow our unified protocol (\S\ref{sec:eval-protocol}).
        \colorbox[RGB]{230,244,244}{Highlighted} rows are our trained models; \textbf{bold} marks the best result within each scale group.
    }
    \label{tab:main-results}
    \centering
    \small
    \newcommand{\toolyes}{\textcolor{green!55!black}{\ding{51}}}
    \newcommand{\toolno}{\textcolor{red!70!black}{\ding{55}}}
    \setlength{\tabcolsep}{4pt}
    \begin{tabular}{l c c c c c c c}
    \toprule
    Model & Tool & AIME25 & HMMT25 & \shortstack{Beyond\\AIME} & \shortstack{IMO\\AnswerBench} & APEX25 & Avg. \\
    \midrule
    \multicolumn{8}{l}{\emph{$<$\!10B scale}} \\
    Qwen3-4B-Thinking-2507 & \toolno & 82.5 & 68.8 & 54.3 & 57.0 & 2.8 & 58.2 \\
    Qwen3.5-4B & \toolno & 75.8 & 72.9 & 58.8 & 59.5 & 0.0 & 60.6 \\
    Qwen3.5-9B & \toolno & 85.8 & 82.1 & 67.3 & 65.0 & 0.0 & 67.2 \\
    \rowcolor[RGB]{230,244,244}
    \trice-4B & \toolno & 79.2 & 71.3 & 58.5 & 61.0 & 5.6 & 61.7 \\
    ASTER-4B$^\dagger$ & \toolyes & 90.0 & 77.1 & 61.7 & -- & -- & -- \\
    AgentMath-8B$^\dagger$ & \toolyes & 84.7 & 71.3 & -- & -- & -- & -- \\
    \rowcolor[RGB]{230,244,244}
    \trice-4B & \toolyes & \textbf{96.7} & \textbf{86.7} & \textbf{71.3} & \textbf{68.9} & \textbf{13.9} & \textbf{72.2} \\
    \midrule
    \multicolumn{8}{l}{\emph{$\sim$\!30B scale}} \\
    Qwen3-30B-A3B-Instruct-2507 & \toolno & 67.5 & 55.8 & 51.3 & 52.3 & 2.8 & 52.5\\
    Qwen3-30B-A3B-Thinking-2507 & \toolno & 88.8 & 75.6 & 65.9 & 66.1 & 0.0 & 67.1 \\
    Qwen3.5-35B-A3B & \toolno & 94.2 & 85.8 & 72.5 & 73.8 & 0.0 & 74.7 \\
    \rowcolor[RGB]{230,244,244}
    \trice-30B & \toolno & 89.2 & 81.7 & 71.0 & 72.3 & 0.0 & 72.8 \\
    GPT-OSS-20B & \toolyes & 86.7 & 83.3 & 63.0 & 60.0 & 8.3 & 63.4 \\
    GLM-4.7-Flash & \toolyes & 95.0 & 84.2 & 76.0 & 68.3 & 11.1 & 71.6 \\
    Nemotron-3-Nano-30B-A3B & \toolyes & 96.7 & 90.4 & 80.0 & 77.0 & 11.1 & 78.8 \\
    AgentMath-30B-A3B$^\dagger$ & \toolyes & 86.4 & 73.8 & -- & -- & -- & -- \\
    \rowcolor[RGB]{230,244,244}
    GLM-4.7-Flash w/ recipe & \toolyes & 98.3 & 89.6 & 81.0 & 78.8 & 13.9 & 80.3\\
    \rowcolor[RGB]{230,244,244}
    \trice-30B & \toolyes & \textbf{99.2} & \textbf{92.5} & \textbf{82.5} & \textbf{80.3} & \textbf{16.7} & \textbf{81.9} \\
    \bottomrule
    \end{tabular}
\end{table*}

\begin{wrapfigure}{r}{0.43\linewidth}
\vspace{-1.0em}
\centering
\includegraphics[width=\linewidth]{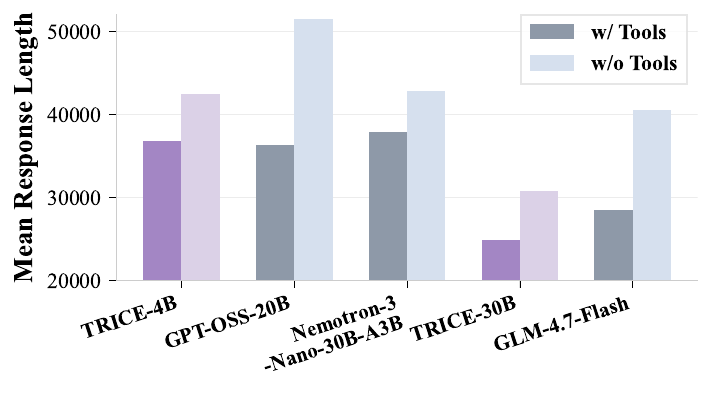}
\caption{TIR token efficiency.}
\label{fig:analysis-token-efficiency}
\vspace{-1.0em}
\end{wrapfigure}

\textbf{Performance.} As shown in Table~\ref{tab:main-results}, our models achieve state-of-the-art tool-integrated reasoning performance, surpassing both existing TIR methods and frontier open-source reasoning models at the same or even larger parameter scales.
Notably, our models yield substantial gains over the base thinking models, specifically, for the 4B and 30B models, we achieve 14.0\% and 14.8\% average improvements, respectively.
It is also noteworthy that on APEX\,2025, a collection of national and international Olympiad problems where most open-source models score near zero, \trice-4B and \trice-30B models reach 13.9\% and 16.7\%, respectively. 
Moreover, our models retain or even improve the text-only reasoning ability on most benchmarks as presented in Table~\ref{tab:main-results}, confirming that with our full-pipeline recipe, the TIR capability is injected without degrading intrinsic reasoning mode.

\textbf{Efficiency.} Beyond solving more problems, tool-integrated reasoning consistently yields shorter responses than text-only 
(Figure~\ref{fig:analysis-token-efficiency}), as code compresses lengthy manual arithmetic into concise executable snippets~\citep{retool, cort}. Also, while achieving superior accuracy, \trice-30B produces substantially shorter responses than same-scale TIR models such as Nemotron-3-Nano and GLM-4.7-Flash.

\subsection{What TIR Unlocks}

\begin{table}[t]
    \caption{Performance (\%) with larger text-only models on representative benchmarks.}
      \vspace{1mm}
    \label{tab:tir-vs-large-text}
    \centering
    \small
    \setlength{\tabcolsep}{4pt}
    \begin{tabular}{l c c c c}
        \toprule
        Model & Tool & HMMT25 & BeyondAIME & IMOAnswerBench \\
        \midrule
        Qwen3-235B-A22B-Thinking & \textcolor{red!70!black}{\ding{55}} & 88.8 & 71.8 & 73.8 \\
        DeepSeek-V3.2-Thinking & \textcolor{red!70!black}{\ding{55}} & 90.8 & 76.8 & 75.0 \\
        \rowcolor[RGB]{230,244,244}
        \trice-30B & \textcolor{green!55!black}{\ding{51}} & \textbf{92.5} & \textbf{82.5} & \textbf{80.3} \\
        \bottomrule
    \end{tabular}
    \vspace{-10pt}
\end{table}

As shown in Table~\ref{tab:tir-vs-large-text}, \trice \ with tools surpasses substantially larger text-only reasoning models. We further analyze what TIR unlocks that parameter scaling alone cannot replicate.

\begin{wrapfigure}{r}{0.35\linewidth}
\vspace{-2.0em}
\centering
\includegraphics[width=\linewidth]{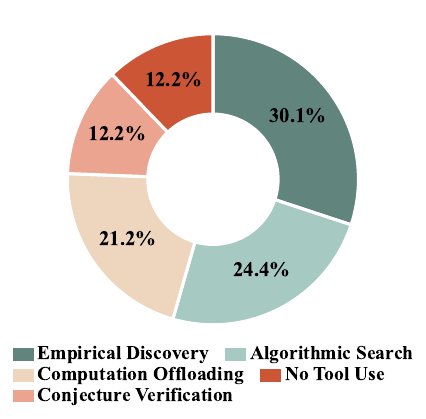}
\caption{Distribution of the roles code-executor plays in \trice-30B trajectories on solved questions.}
\label{fig:analysis-purpose-pie}
\vspace{-1.0em}
\end{wrapfigure}

\textbf{1) Code executor is not merely a calculator, but a cognitive tool.}
We annotate each solved trajectory with a primary computational purpose with Gemini-3-Flash: empirical discovery, algorithmic search, computation offloading, or conjecture verification (definitions in Appendix~\ref{app:code-use-analysis}). As Figure~\ref{fig:analysis-purpose-pie} shows, code is more often used to \emph{explore and discover}, such as computing small instances to conjecture a pattern, or enumerating a combinatorial space instead of merely performing precise arithmetic. This matches the nature of competition-level questions, which require identifying structure from concrete cases or searching over configurations that grow combinatorially.

\textbf{2) Scaling text-only reasoning is not a silver bullet.}
TIR uniquely solves problems dominated by code-critical algorithmic search. These problems remain challenging for text-only scaling, since natural-language simulation of iterative computation is both context-inefficient and error-prone, especially for large-scale search and multi-case exploration~\cite{understandingtir}.

\subsection{Generalization Analysis}

\begin{table}[t]
    \caption{Generalization performance (\%) to other domains.}
      \vspace{1mm}
    \label{tab:domain-generalization}
    \centering
    \small
    \setlength{\tabcolsep}{5pt}
    \newcommand{\baseacc}[1]{\makebox[2.4em][r]{#1}\makebox[4.1em][l]{}}
    \newcommand{\tiracc}[2]{\makebox[2.4em][r]{\textbf{#1}}\makebox[4.1em][l]{~\textcolor{green!55!black}{(#2)}}}
    \begin{tabular}{l c c c c}
      \toprule
      Model & Tool & FrontierScience & GPQA-Diamond & LiveCodeBench \\
      \midrule
      Qwen3-4B-Thinking-2507 & \textcolor{red!70!black}{\ding{55}} & \baseacc{27.5} & \baseacc{64.4} & \baseacc{51.4} \\
      \trice-4B & \textcolor{green!55!black}{\ding{51}} &
        \tiracc{42.0}{+14.5} &
        \tiracc{68.8}{+4.4} &
        \tiracc{55.6}{+4.2} \\
      \midrule
      Qwen3-30B-A3B-Thinking-2507 & \textcolor{red!70!black}{\ding{55}} & \baseacc{44.9} & \baseacc{71.2} & \baseacc{61.5} \\
      \trice-30B & \textcolor{green!55!black}{\ding{51}} &
        \tiracc{53.0}{+8.1} &
        \tiracc{75.4}{+4.2} &
        \tiracc{73.2}{+11.7} \\
      \bottomrule
    \end{tabular}
    \vspace{-12pt}
\end{table}

\textbf{Cross-domain transfer.}
Although trained only on math data, the learned interleaved reasoning pattern transfers to different domains and tasks. As shown in Table~\ref{tab:domain-generalization}, across diverse benchmarks, including FrontierScience~\citep{wang2026frontierscience}, which requires scientific computation, LiveCodeBench~\citep{jain2024livecodebench}, which evaluates code generation ability, and the knowledge-intensive GPQA-Diamond~\citep{rein2024gpqa}, our models consistently achieve non-trivial improvements over the base models, with gains of up to 14.5\%.

\textbf{Cross-model transfer.}
Our recipe is not specific to a single model family. We apply the recipe to GLM-4.7-Flash, a model that already possesses native TIR capability~\cite{5team2025glm45agenticreasoningcoding}. As shown in Table~\ref{tab:main-results}, GLM-4.7-Flash w/ recipe further improves over the original GLM-4.7-Flash on most benchmarks, confirming that our data and training recipe provide complementary signal even for models with existing tool-use ability. Full training details are in Appendix~\ref{app:rl-training-dynamics}.

\section{Conclusion}
\label{sec:conclusion}
{In this study, we addressed the challenge of teaching strong thinking models to perform tool-integrated reasoning. We developed a comprehensive training pipeline, spanning from data curation to SFT and RL training recipe. Through our approaches, our models achieved SOTA performance in TIR while preserving no-tool capabilities and further generalizing to different tasks. We hope this study provides useful insights for the community to train future reasoning models.}

\bibliographystyle{plainnat}
\bibliography{custom}

@misc{zhang2026aster,
      title={ASTER: Agentic Scaling with Tool-integrated Extended Reasoning}, 
      author={Xuqin Zhang and Quan He and Zhenrui Zheng and Zongzhang Zhang and Xu He and Dong Li},
      year={2026},
      eprint={2602.01204},
      archivePrefix={arXiv},
      primaryClass={cs.CL},
      url={https://arxiv.org/abs/2602.01204}, 
}

@misc{zhang2026onpolicyrlmeetsoffpolicy,
      title={On-Policy RL Meets Off-Policy Experts: Harmonizing Supervised Fine-Tuning and Reinforcement Learning via Dynamic Weighting}, 
      author={Wenhao Zhang and Yuexiang Xie and Yuchang Sun and Yanxi Chen and Guoyin Wang and Yaliang Li and Bolin Ding and Jingren Zhou},
      year={2026},
      eprint={2508.11408},
      archivePrefix={arXiv},
      primaryClass={cs.LG},
      url={https://arxiv.org/abs/2508.11408}, 
}

@misc{simpletir,
      title={SimpleTIR: End-to-End Reinforcement Learning for Multi-Turn Tool-Integrated Reasoning}, 
      author={Zhenghai Xue and Longtao Zheng and Qian Liu and Yingru Li and Xiaosen Zheng and Zejun Ma and Bo An},
      year={2025},
      eprint={2509.02479},
      archivePrefix={arXiv},
      primaryClass={cs.LG},
      url={https://arxiv.org/abs/2509.02479}, 
}

@misc{ma2025stabilizing,
      title={Stabilizing MoE Reinforcement Learning by Aligning Training and Inference Routers}, 
      author={Wenhan Ma and Hailin Zhang and Liang Zhao and Yifan Song and Yudong Wang and Zhifang Sui and Fuli Luo},
      year={2025},
      eprint={2510.11370},
      archivePrefix={arXiv},
      primaryClass={cs.CL},
      url={https://arxiv.org/abs/2510.11370}, 
}

@inproceedings{luo2025through,
    title = "Through the Valley: Path to Effective Long {C}o{T} Training for Small Language Models",
    author = "Luo, Renjie  and
      Li, Jiaxi  and
      Huang, Chen  and
      Lu, Wei",
    editor = "Christodoulopoulos, Christos  and
      Chakraborty, Tanmoy  and
      Rose, Carolyn  and
      Peng, Violet",
    booktitle = "Proceedings of the 2025 Conference on Empirical Methods in Natural Language Processing",
    month = nov,
    year = "2025",
    address = "Suzhou, China",
    publisher = "Association for Computational Linguistics",
    url = "https://aclanthology.org/2025.emnlp-main.251/",
    doi = "10.18653/v1/2025.emnlp-main.251",
    pages = "4972--4992",
    ISBN = "979-8-89176-332-6"
}

@misc{understandingtir,
      title={Understanding Tool-Integrated Reasoning}, 
      author={Heng Lin and Zhongwen Xu},
      year={2025},
      eprint={2508.19201},
      archivePrefix={arXiv},
      primaryClass={cs.LG},
      url={https://arxiv.org/abs/2508.19201}, 
}

@misc{li2026discovery,
      title={Discovery and Reinforcement of Tool-Integrated Reasoning Chains via Rollout Trees}, 
      author={Kun Li and Zenan Xu and Junan Li and Zengrui Jin and Jinghao Deng and Zexuan Qiu and Bo Zhou},
      year={2026},
      eprint={2601.08274},
      archivePrefix={arXiv},
      primaryClass={cs.CL},
      url={https://arxiv.org/abs/2601.08274}, 
}

@misc{cort,
      title={CoRT: Code-integrated Reasoning within Thinking}, 
      author={Chengpeng Li and Zhengyang Tang and Ziniu Li and Mingfeng Xue and Keqin Bao and Tian Ding and Ruoyu Sun and Benyou Wang and Xiang Wang and Junyang Lin and Dayiheng Liu},
      year={2025},
      eprint={2506.09820},
      archivePrefix={arXiv},
      primaryClass={cs.CL},
      url={https://arxiv.org/abs/2506.09820}, 
}

@misc{retool,
      title={ReTool: Reinforcement Learning for Strategic Tool Use in LLMs}, 
      author={Jiazhan Feng and Shijue Huang and Xingwei Qu and Ge Zhang and Yujia Qin and Baoquan Zhong and Chengquan Jiang and Jinxin Chi and Wanjun Zhong},
      year={2025},
      eprint={2504.11536},
      archivePrefix={arXiv},
      primaryClass={cs.CL},
      url={https://arxiv.org/abs/2504.11536}, 
}

@InProceedings{pal,
  title = 	 {{PAL}: Program-aided Language Models},
  author =       {Gao, Luyu and Madaan, Aman and Zhou, Shuyan and Alon, Uri and Liu, Pengfei and Yang, Yiming and Callan, Jamie and Neubig, Graham},
  booktitle = 	 {Proceedings of the 40th International Conference on Machine Learning},
  pages = 	 {10764--10799},
  year = 	 {2023},
  editor = 	 {Krause, Andreas and Brunskill, Emma and Cho, Kyunghyun and Engelhardt, Barbara and Sabato, Sivan and Scarlett, Jonathan},
  volume = 	 {202},
  series = 	 {Proceedings of Machine Learning Research},
  month = 	 {23--29 Jul},
  publisher =    {PMLR},
  url = 	 {https://proceedings.mlr.press/v202/gao23f.html}
}

@book{sutton2018reinforcement,
  title={Reinforcement learning: An introduction},
  author={Sutton, Richard S and Barto, Andrew G},
  year={2018},
  publisher={MIT press}
}

@misc{deepseekai2025deepseekr1incentivizingreasoningcapability,
      title={DeepSeek-R1: Incentivizing Reasoning Capability in LLMs via Reinforcement Learning}, 
      author={DeepSeek-AI and Daya Guo and Dejian Yang and Haowei Zhang and Junxiao Song and Ruoyu Zhang and Runxin Xu and Qihao Zhu and Shirong Ma and Peiyi Wang and Xiao Bi and Xiaokang Zhang and Xingkai Yu and Yu Wu and Z. F. Wu and Zhibin Gou and Zhihong Shao and Zhuoshu Li and Ziyi Gao and Aixin Liu and Bing Xue and Bingxuan Wang and Bochao Wu and Bei Feng and Chengda Lu and Chenggang Zhao and Chengqi Deng and Chenyu Zhang and Chong Ruan and Damai Dai and Deli Chen and Dongjie Ji and Erhang Li and Fangyun Lin and Fucong Dai and Fuli Luo and Guangbo Hao and Guanting Chen and Guowei Li and H. Zhang and Han Bao and Hanwei Xu and Haocheng Wang and Honghui Ding and Huajian Xin and Huazuo Gao and Hui Qu and Hui Li and Jianzhong Guo and Jiashi Li and Jiawei Wang and Jingchang Chen and Jingyang Yuan and Junjie Qiu and Junlong Li and J. L. Cai and Jiaqi Ni and Jian Liang and Jin Chen and Kai Dong and Kai Hu and Kaige Gao and Kang Guan and Kexin Huang and Kuai Yu and Lean Wang and Lecong Zhang and Liang Zhao and Litong Wang and Liyue Zhang and Lei Xu and Leyi Xia and Mingchuan Zhang and Minghua Zhang and Minghui Tang and Meng Li and Miaojun Wang and Mingming Li and Ning Tian and Panpan Huang and Peng Zhang and Qiancheng Wang and Qinyu Chen and Qiushi Du and Ruiqi Ge and Ruisong Zhang and Ruizhe Pan and Runji Wang and R. J. Chen and R. L. Jin and Ruyi Chen and Shanghao Lu and Shangyan Zhou and Shanhuang Chen and Shengfeng Ye and Shiyu Wang and Shuiping Yu and Shunfeng Zhou and Shuting Pan and S. S. Li and Shuang Zhou and Shaoqing Wu and Shengfeng Ye and Tao Yun and Tian Pei and Tianyu Sun and T. Wang and Wangding Zeng and Wanjia Zhao and Wen Liu and Wenfeng Liang and Wenjun Gao and Wenqin Yu and Wentao Zhang and W. L. Xiao and Wei An and Xiaodong Liu and Xiaohan Wang and Xiaokang Chen and Xiaotao Nie and Xin Cheng and Xin Liu and Xin Xie and Xingchao Liu and Xinyu Yang and Xinyuan Li and Xuecheng Su and Xuheng Lin and X. Q. Li and Xiangyue Jin and Xiaojin Shen and Xiaosha Chen and Xiaowen Sun and Xiaoxiang Wang and Xinnan Song and Xinyi Zhou and Xianzu Wang and Xinxia Shan and Y. K. Li and Y. Q. Wang and Y. X. Wei and Yang Zhang and Yanhong Xu and Yao Li and Yao Zhao and Yaofeng Sun and Yaohui Wang and Yi Yu and Yichao Zhang and Yifan Shi and Yiliang Xiong and Ying He and Yishi Piao and Yisong Wang and Yixuan Tan and Yiyang Ma and Yiyuan Liu and Yongqiang Guo and Yuan Ou and Yuduan Wang and Yue Gong and Yuheng Zou and Yujia He and Yunfan Xiong and Yuxiang Luo and Yuxiang You and Yuxuan Liu and Yuyang Zhou and Y. X. Zhu and Yanhong Xu and Yanping Huang and Yaohui Li and Yi Zheng and Yuchen Zhu and Yunxian Ma and Ying Tang and Yukun Zha and Yuting Yan and Z. Z. Ren and Zehui Ren and Zhangli Sha and Zhe Fu and Zhean Xu and Zhenda Xie and Zhengyan Zhang and Zhewen Hao and Zhicheng Ma and Zhigang Yan and Zhiyu Wu and Zihui Gu and Zijia Zhu and Zijun Liu and Zilin Li and Ziwei Xie and Ziyang Song and Zizheng Pan and Zhen Huang and Zhipeng Xu and Zhongyu Zhang and Zhen Zhang},
      year={2025},
      eprint={2501.12948},
      archivePrefix={arXiv},
      primaryClass={cs.CL},
      url={https://arxiv.org/abs/2501.12948}, 
}

@misc{openai2024gpt4technicalreport,
      title={GPT-4 Technical Report}, 
      author={OpenAI and Josh Achiam and Steven Adler and Sandhini Agarwal and Lama Ahmad and Ilge Akkaya and Florencia Leoni Aleman and Diogo Almeida and Janko Altenschmidt and Sam Altman and Shyamal Anadkat and Red Avila and Igor Babuschkin and Suchir Balaji and Valerie Balcom and Paul Baltescu and Haiming Bao and Mohammad Bavarian and Jeff Belgum and Irwan Bello and Jake Berdine and Gabriel Bernadett-Shapiro and Christopher Berner and Lenny Bogdonoff and Oleg Boiko and Madelaine Boyd and Anna-Luisa Brakman and Greg Brockman and Tim Brooks and Miles Brundage and Kevin Button and Trevor Cai and Rosie Campbell and Andrew Cann and Brittany Carey and Chelsea Carlson and Rory Carmichael and Brooke Chan and Che Chang and Fotis Chantzis and Derek Chen and Sully Chen and Ruby Chen and Jason Chen and Mark Chen and Ben Chess and Chester Cho and Casey Chu and Hyung Won Chung and Dave Cummings and Jeremiah Currier and Yunxing Dai and Cory Decareaux and Thomas Degry and Noah Deutsch and Damien Deville and Arka Dhar and David Dohan and Steve Dowling and Sheila Dunning and Adrien Ecoffet and Atty Eleti and Tyna Eloundou and David Farhi and Liam Fedus and Niko Felix and Simón Posada Fishman and Juston Forte and Isabella Fulford and Leo Gao and Elie Georges and Christian Gibson and Vik Goel and Tarun Gogineni and Gabriel Goh and Rapha Gontijo-Lopes and Jonathan Gordon and Morgan Grafstein and Scott Gray and Ryan Greene and Joshua Gross and Shixiang Shane Gu and Yufei Guo and Chris Hallacy and Jesse Han and Jeff Harris and Yuchen He and Mike Heaton and Johannes Heidecke and Chris Hesse and Alan Hickey and Wade Hickey and Peter Hoeschele and Brandon Houghton and Kenny Hsu and Shengli Hu and Xin Hu and Joost Huizinga and Shantanu Jain and Shawn Jain and Joanne Jang and Angela Jiang and Roger Jiang and Haozhun Jin and Denny Jin and Shino Jomoto and Billie Jonn and Heewoo Jun and Tomer Kaftan and Łukasz Kaiser and Ali Kamali and Ingmar Kanitscheider and Nitish Shirish Keskar and Tabarak Khan and Logan Kilpatrick and Jong Wook Kim and Christina Kim and Yongjik Kim and Jan Hendrik Kirchner and Jamie Kiros and Matt Knight and Daniel Kokotajlo and Łukasz Kondraciuk and Andrew Kondrich and Aris Konstantinidis and Kyle Kosic and Gretchen Krueger and Vishal Kuo and Michael Lampe and Ikai Lan and Teddy Lee and Jan Leike and Jade Leung and Daniel Levy and Chak Ming Li and Rachel Lim and Molly Lin and Stephanie Lin and Mateusz Litwin and Theresa Lopez and Ryan Lowe and Patricia Lue and Anna Makanju and Kim Malfacini and Sam Manning and Todor Markov and Yaniv Markovski and Bianca Martin and Katie Mayer and Andrew Mayne and Bob McGrew and Scott Mayer McKinney and Christine McLeavey and Paul McMillan and Jake McNeil and David Medina and Aalok Mehta and Jacob Menick and Luke Metz and Andrey Mishchenko and Pamela Mishkin and Vinnie Monaco and Evan Morikawa and Daniel Mossing and Tong Mu and Mira Murati and Oleg Murk and David Mély and Ashvin Nair and Reiichiro Nakano and Rajeev Nayak and Arvind Neelakantan and Richard Ngo and Hyeonwoo Noh and Long Ouyang and Cullen O'Keefe and Jakub Pachocki and Alex Paino and Joe Palermo and Ashley Pantuliano and Giambattista Parascandolo and Joel Parish and Emy Parparita and Alex Passos and Mikhail Pavlov and Andrew Peng and Adam Perelman and Filipe de Avila Belbute Peres and Michael Petrov and Henrique Ponde de Oliveira Pinto and Michael and Pokorny and Michelle Pokrass and Vitchyr H. Pong and Tolly Powell and Alethea Power and Boris Power and Elizabeth Proehl and Raul Puri and Alec Radford and Jack Rae and Aditya Ramesh and Cameron Raymond and Francis Real and Kendra Rimbach and Carl Ross and Bob Rotsted and Henri Roussez and Nick Ryder and Mario Saltarelli and Ted Sanders and Shibani Santurkar and Girish Sastry and Heather Schmidt and David Schnurr and John Schulman and Daniel Selsam and Kyla Sheppard and Toki Sherbakov and Jessica Shieh and Sarah Shoker and Pranav Shyam and Szymon Sidor and Eric Sigler and Maddie Simens and Jordan Sitkin and Katarina Slama and Ian Sohl and Benjamin Sokolowsky and Yang Song and Natalie Staudacher and Felipe Petroski Such and Natalie Summers and Ilya Sutskever and Jie Tang and Nikolas Tezak and Madeleine B. Thompson and Phil Tillet and Amin Tootoonchian and Elizabeth Tseng and Preston Tuggle and Nick Turley and Jerry Tworek and Juan Felipe Cerón Uribe and Andrea Vallone and Arun Vijayvergiya and Chelsea Voss and Carroll Wainwright and Justin Jay Wang and Alvin Wang and Ben Wang and Jonathan Ward and Jason Wei and CJ Weinmann and Akila Welihinda and Peter Welinder and Jiayi Weng and Lilian Weng and Matt Wiethoff and Dave Willner and Clemens Winter and Samuel Wolrich and Hannah Wong and Lauren Workman and Sherwin Wu and Jeff Wu and Michael Wu and Kai Xiao and Tao Xu and Sarah Yoo and Kevin Yu and Qiming Yuan and Wojciech Zaremba and Rowan Zellers and Chong Zhang and Marvin Zhang and Shengjia Zhao and Tianhao Zheng and Juntang Zhuang and William Zhuk and Barret Zoph},
      year={2024},
      eprint={2303.08774},
      archivePrefix={arXiv},
      primaryClass={cs.CL},
      url={https://arxiv.org/abs/2303.08774}, 
}

@misc{jain2024livecodebench,
      title={LiveCodeBench: Holistic and Contamination Free Evaluation of Large Language Models for Code}, 
      author={Naman Jain and King Han and Alex Gu and Wen-Ding Li and Fanjia Yan and Tianjun Zhang and Sida Wang and Armando Solar-Lezama and Koushik Sen and Ion Stoica},
      year={2024},
      eprint={2403.07974},
      archivePrefix={arXiv},
      primaryClass={cs.SE},
      url={https://arxiv.org/abs/2403.07974}, 
}

@misc{deepseekmath,
      title={DeepSeekMath: Pushing the Limits of Mathematical Reasoning in Open Language Models}, 
      author={Zhihong Shao and Peiyi Wang and Qihao Zhu and Runxin Xu and Junxiao Song and Xiao Bi and Haowei Zhang and Mingchuan Zhang and Y. K. Li and Y. Wu and Daya Guo},
      year={2024},
      eprint={2402.03300},
      archivePrefix={arXiv},
      primaryClass={cs.CL},
      url={https://arxiv.org/abs/2402.03300}, 
}

@inproceedings{rein2024gpqa,
  title={Gpqa: A graduate-level google-proof q\&a benchmark},
  author={Rein, David and Hou, Betty Li and Stickland, Asa Cooper and Petty, Jackson and Pang, Richard Yuanzhe and Dirani, Julien and Michael, Julian and Bowman, Samuel R},
  booktitle={First Conference on Language Modeling},
  year={2024}
}

@misc{zuo2025ttrl,
      title={TTRL: Test-Time Reinforcement Learning}, 
      author={Yuxin Zuo and Kaiyan Zhang and Li Sheng and Shang Qu and Ganqu Cui and Xuekai Zhu and Haozhan Li and Yuchen Zhang and Xinwei Long and Ermo Hua and Biqing Qi and Youbang Sun and Zhiyuan Ma and Lifan Yuan and Ning Ding and Bowen Zhou},
      year={2025},
      eprint={2504.16084},
      archivePrefix={arXiv},
      primaryClass={cs.CL},
      url={https://arxiv.org/abs/2504.16084}, 
}

@misc{zeng2025simplerlzooinvestigatingtamingzero,
      title={SimpleRL-Zoo: Investigating and Taming Zero Reinforcement Learning for Open Base Models in the Wild}, 
      author={Weihao Zeng and Yuzhen Huang and Qian Liu and Wei Liu and Keqing He and Zejun Ma and Junxian He},
      year={2025},
      eprint={2503.18892},
      archivePrefix={arXiv},
      primaryClass={cs.LG},
      url={https://arxiv.org/abs/2503.18892}, 
}

@article{zhang2025tool,
  title={Tool-R1: Sample-Efficient Reinforcement Learning for Agentic Tool Use},
  author={Zhang, Yabo and Zeng, Yihan and Li, Qingyun and Hu, Zhen and Han, Kavin and Zuo, Wangmeng},
  journal={arXiv preprint arXiv:2509.12867},
  year={2025}
}

@article{yu2025demystifying,
  title={Demystifying reinforcement learning in agentic reasoning},
  author={Yu, Zhaochen and Yang, Ling and Zou, Jiaru and Yan, Shuicheng and Wang, Mengdi},
  journal={arXiv preprint arXiv:2510.11701},
  year={2025}
}

@article{du2025nemotron,
  title={Nemotron-Math: Efficient Long-Context Distillation of Mathematical Reasoning from Multi-Mode Supervision},
  author={Du, Wei and Toshniwal, Shubham and Kisacanin, Branislav and Mahdavi, Sadegh and Moshkov, Ivan and Armstrong, George and Ge, Stephen and Minasyan, Edgar and Chen, Feng and Gitman, Igor},
  journal={arXiv preprint arXiv:2512.15489},
  year={2025}
}

@article{dong2025agentic,
  title={Agentic reinforced policy optimization},
  author={Dong, Guanting and Mao, Hangyu and Ma, Kai and Bao, Licheng and Chen, Yifei and Wang, Zhongyuan and Chen, Zhongxia and Du, Jiazhen and Wang, Huiyang and Zhang, Fuzheng and others},
  journal={arXiv preprint arXiv:2507.19849},
  year={2025}
}

@article{wang2025information,
  title={Information Gain-based Policy Optimization: A Simple and Effective Approach for Multi-Turn LLM Agents},
  author={Wang, Guoqing and Dai, Sunhao and Ye, Guangze and Gan, Zeyu and Yao, Wei and Deng, Yong and Wu, Xiaofeng and Ying, Zhenzhe},
  journal={arXiv preprint arXiv:2510.14967},
  year={2025}
}

@article{du2025generalizable,
  title={Generalizable end-to-end tool-use rl with synthetic codegym},
  author={Du, Weihua and Gong, Hailei and Ling, Zhan and Liu, Kang and Shen, Lingfeng and Yao, Xuesong and Xu, Yufei and Shi, Dingyuan and Yang, Yiming and Chen, Jiecao},
  journal={arXiv preprint arXiv:2509.17325},
  year={2025}
}

@article{shang2025rstar2,
  title={rstar2-agent: Agentic reasoning technical report},
  author={Shang, Ning and Liu, Yifei and Zhu, Yi and Zhang, Li Lyna and Xu, Weijiang and Guan, Xinyu and Zhang, Buze and Dong, Bingcheng and Zhou, Xudong and Zhang, Bowen and others},
  journal={arXiv preprint arXiv:2508.20722},
  year={2025}
}

@article{xia2025agent0,
  title={Agent0: Unleashing self-evolving agents from zero data via tool-integrated reasoning},
  author={Xia, Peng and Zeng, Kaide and Liu, Jiaqi and Qin, Can and Wu, Fang and Zhou, Yiyang and Xiong, Caiming and Yao, Huaxiu},
  journal={arXiv preprint arXiv:2511.16043},
  year={2025}
}

@article{cui2025process,
  title={Process reinforcement through implicit rewards},
  author={Cui, Ganqu and Yuan, Lifan and Wang, Zefan and Wang, Hanbin and Zhang, Yuchen and Chen, Jiacheng and Li, Wendi and He, Bingxiang and Fan, Yuchen and Yu, Tianyu and others},
  journal={arXiv preprint arXiv:2502.01456},
  year={2025}
}

@article{cui2025entropy,
  title={The entropy mechanism of reinforcement learning for reasoning language models},
  author={Cui, Ganqu and Zhang, Yuchen and Chen, Jiacheng and Yuan, Lifan and Wang, Zhi and Zuo, Yuxin and Li, Haozhan and Fan, Yuchen and Chen, Huayu and Chen, Weize and others},
  journal={arXiv preprint arXiv:2505.22617},
  year={2025}
}

@article{zhang2025survey,
  title={A survey of reinforcement learning for large reasoning models},
  author={Zhang, Kaiyan and Zuo, Yuxin and He, Bingxiang and Sun, Youbang and Liu, Runze and Jiang, Che and Fan, Yuchen and Tian, Kai and Jia, Guoli and Li, Pengfei and others},
  journal={arXiv preprint arXiv:2509.08827},
  year={2025}
}

@article{yuan2025f,
  title={From $ f (x) $ and $ g (x) $ to $ f (g (x)) $: LLMs Learn New Skills in RL by Composing Old Ones},
  author={Yuan, Lifan and Chen, Weize and Zhang, Yuchen and Cui, Ganqu and Wang, Hanbin and You, Ziming and Ding, Ning and Liu, Zhiyuan and Sun, Maosong and Peng, Hao},
  journal={arXiv preprint arXiv:2509.25123},
  year={2025}
}

@article{chen2025p1,
  title={P1: Mastering Physics Olympiads with Reinforcement Learning},
  author={Chen, Jiacheng and Cheng, Qianjia and Yu, Fangchen and Wan, Haiyuan and Zhang, Yuchen and Zheng, Shenghe and Yao, Junchi and Zhang, Qingyang and He, Haonan and Luo, Yun and others},
  journal={arXiv preprint arXiv:2511.13612},
  year={2025}
}

@article{luo2026p1,
  title={P1-VL: bridging visual perception and scientific reasoning in physics Olympiads},
  author={Luo, Yun and Wang, Futing and Cheng, Qianjia and Yu, Fangchen and Lei, Haodi and Yan, Jianhao and Li, Chenxi and Chen, Jiacheng and Zhao, Yufeng and Wan, Haiyuan and others},
  journal={arXiv preprint arXiv:2602.09443},
  year={2026}
}

@article{guha2025openthoughts,
  title={Openthoughts: Data recipes for reasoning models},
  author={Guha, Etash and Marten, Ryan and Keh, Sedrick and Raoof, Negin and Smyrnis, Georgios and Bansal, Hritik and Nezhurina, Marianna and Mercat, Jean and Vu, Trung and Sprague, Zayne and others},
  journal={arXiv preprint arXiv:2506.04178},
  year={2025}
}

@inproceedings{jiang2025rocode,
  title={ROCODE: Integrating backtracking mechanism and program analysis in large language models for code generation},
  author={Jiang, Xue and Dong, Yihong and Tao, Yongding and Liu, Huanyu and Jin, Zhi and Li, Ge},
  booktitle={2025 IEEE/ACM 47th International Conference on Software Engineering (ICSE)},
  pages={334--346},
  year={2025},
  organization={IEEE}
}

@inproceedings{gao2025train,
  title={How to train long-context language models (effectively)},
  author={Gao, Tianyu and Wettig, Alexander and Yen, Howard and Chen, Danqi},
  booktitle={Proceedings of the 63rd Annual Meeting of the Association for Computational Linguistics (Volume 1: Long Papers)},
  pages={7376--7399},
  year={2025}
}

@article{he2025justrl,
  title={Justrl: Scaling a 1.5 b llm with a simple rl recipe},
  author={He, Bingxiang and Qu, Zekai and Liu, Zeyuan and Chen, Yinghao and Zuo, Yuxin and Qian, Cheng and Zhang, Kaiyan and Chen, Weize and Xiao, Chaojun and Cui, Ganqu and others},
  journal={arXiv preprint arXiv:2512.16649},
  year={2025}
}

@article{minimax2026m,
    author = {Minimax Team},
    title = {MiniMax M2.7: Early Echoes of Self-Evolution},
    journal = {https://www.minimax.io/news/minimax-m27-en},
    year = {2026}
}

@article{zheng2025stabilizing,
  title={Stabilizing reinforcement learning with llms: Formulation and practices},
  author={Zheng, Chujie and Dang, Kai and Yu, Bowen and Li, Mingze and Jiang, Huiqiang and Lin, Junrong and Liu, Yuqiong and Lin, Hao and Wu, Chencan and Hu, Feng and others},
  journal={arXiv preprint arXiv:2512.01374},
  year={2025}
}

@article{liu2025speed,
  title={When speed kills stability: Demystifying rl collapse from the training-inference mismatch},
  author={Liu, Jiacai and Li, Yingru and Fu, Yuqian and Wang, Jiawei and Liu, Qian and Shen, Yu},
  journal={Notion Blog},
  year={2025}
}

@article{li2025taming,
  title={Taming the Tail: Stable LLM Reinforcement Learning via Dynamic Vocabulary Pruning},
  author={Li, Yingru and Xu, Jiawei and Liu, Jiacai and Tong, Yuxuan and Li, Ziniu and Cai, Tianle and Zhang, Ge and Liu, Qian and Wang, Baoxiang},
  journal={arXiv preprint arXiv:2512.23087},
  year={2025}
}

@misc{yang2025qwen3technicalreport,
      title={Qwen3 Technical Report}, 
      author={An Yang and Anfeng Li and Baosong Yang and Beichen Zhang and Binyuan Hui and Bo Zheng and Bowen Yu and Chang Gao and Chengen Huang and Chenxu Lv and Chujie Zheng and Dayiheng Liu and Fan Zhou and Fei Huang and Feng Hu and Hao Ge and Haoran Wei and Huan Lin and Jialong Tang and Jian Yang and Jianhong Tu and Jianwei Zhang and Jianxin Yang and Jiaxi Yang and Jing Zhou and Jingren Zhou and Junyang Lin and Kai Dang and Keqin Bao and Kexin Yang and Le Yu and Lianghao Deng and Mei Li and Mingfeng Xue and Mingze Li and Pei Zhang and Peng Wang and Qin Zhu and Rui Men and Ruize Gao and Shixuan Liu and Shuang Luo and Tianhao Li and Tianyi Tang and Wenbiao Yin and Xingzhang Ren and Xinyu Wang and Xinyu Zhang and Xuancheng Ren and Yang Fan and Yang Su and Yichang Zhang and Yinger Zhang and Yu Wan and Yuqiong Liu and Zekun Wang and Zeyu Cui and Zhenru Zhang and Zhipeng Zhou and Zihan Qiu},
      year={2025},
      eprint={2505.09388},
      archivePrefix={arXiv},
      primaryClass={cs.CL},
      url={https://arxiv.org/abs/2505.09388}, 
}

@misc{5team2025glm45agenticreasoningcoding,
      title={GLM-4.5: Agentic, Reasoning, and Coding (ARC) Foundation Models}, 
      author={ 5 Team and Aohan Zeng and Xin Lv and Qinkai Zheng and Zhenyu Hou and Bin Chen and Chengxing Xie and Cunxiang Wang and Da Yin and Hao Zeng and Jiajie Zhang and Kedong Wang and Lucen Zhong and Mingdao Liu and Rui Lu and Shulin Cao and Xiaohan Zhang and Xuancheng Huang and Yao Wei and Yean Cheng and Yifan An and Yilin Niu and Yuanhao Wen and Yushi Bai and Zhengxiao Du and Zihan Wang and Zilin Zhu and Bohan Zhang and Bosi Wen and Bowen Wu and Bowen Xu and Can Huang and Casey Zhao and Changpeng Cai and Chao Yu and Chen Li and Chendi Ge and Chenghua Huang and Chenhui Zhang and Chenxi Xu and Chenzheng Zhu and Chuang Li and Congfeng Yin and Daoyan Lin and Dayong Yang and Dazhi Jiang and Ding Ai and Erle Zhu and Fei Wang and Gengzheng Pan and Guo Wang and Hailong Sun and Haitao Li and Haiyang Li and Haiyi Hu and Hanyu Zhang and Hao Peng and Hao Tai and Haoke Zhang and Haoran Wang and Haoyu Yang and He Liu and He Zhao and Hongwei Liu and Hongxi Yan and Huan Liu and Huilong Chen and Ji Li and Jiajing Zhao and Jiamin Ren and Jian Jiao and Jiani Zhao and Jianyang Yan and Jiaqi Wang and Jiayi Gui and Jiayue Zhao and Jie Liu and Jijie Li and Jing Li and Jing Lu and Jingsen Wang and Jingwei Yuan and Jingxuan Li and Jingzhao Du and Jinhua Du and Jinxin Liu and Junkai Zhi and Junli Gao and Ke Wang and Lekang Yang and Liang Xu and Lin Fan and Lindong Wu and Lintao Ding and Lu Wang and Man Zhang and Minghao Li and Minghuan Xu and Mingming Zhao and Mingshu Zhai and Pengfan Du and Qian Dong and Shangde Lei and Shangqing Tu and Shangtong Yang and Shaoyou Lu and Shijie Li and Shuang Li and Shuang-Li and Shuxun Yang and Sibo Yi and Tianshu Yu and Wei Tian and Weihan Wang and Wenbo Yu and Weng Lam Tam and Wenjie Liang and Wentao Liu and Xiao Wang and Xiaohan Jia and Xiaotao Gu and Xiaoying Ling and Xin Wang and Xing Fan and Xingru Pan and Xinyuan Zhang and Xinze Zhang and Xiuqing Fu and Xunkai Zhang and Yabo Xu and Yandong Wu and Yida Lu and Yidong Wang and Yilin Zhou and Yiming Pan and Ying Zhang and Yingli Wang and Yingru Li and Yinpei Su and Yipeng Geng and Yitong Zhu and Yongkun Yang and Yuhang Li and Yuhao Wu and Yujiang Li and Yunan Liu and Yunqing Wang and Yuntao Li and Yuxuan Zhang and Zezhen Liu and Zhen Yang and Zhengda Zhou and Zhongpei Qiao and Zhuoer Feng and Zhuorui Liu and Zichen Zhang and Zihan Wang and Zijun Yao and Zikang Wang and Ziqiang Liu and Ziwei Chai and Zixuan Li and Zuodong Zhao and Wenguang Chen and Jidong Zhai and Bin Xu and Minlie Huang and Hongning Wang and Juanzi Li and Yuxiao Dong and Jie Tang},
      year={2025},
      eprint={2508.06471},
      archivePrefix={arXiv},
      primaryClass={cs.CL},
      url={https://arxiv.org/abs/2508.06471}, 
}

@misc{slime_github,
  author       = {Zilin Zhu and Chengxing Xie and Xin Lv and slime Contributors},
  title        = {slime: An LLM post-training framework for RL Scaling},
  year         = {2025},
  howpublished = {\url{https://github.com/THUDM/slime}},
  note         = {GitHub repository. Corresponding author: Xin Lv},
  urldate      = {2025-06-19}
}

@misc{kimiteam2026kimik25visualagentic,
      title={Kimi K2.5: Visual Agentic Intelligence}, 
      author={Kimi Team and Tongtong Bai and Yifan Bai and Yiping Bao and S. H. Cai and Yuan Cao and Y. Charles and H. S. Che and Cheng Chen and Guanduo Chen and Huarong Chen and Jia Chen and Jiahao Chen and Jianlong Chen and Jun Chen and Kefan Chen and Liang Chen and Ruijue Chen and Xinhao Chen and Yanru Chen and Yanxu Chen and Yicun Chen and Yimin Chen and Yingjiang Chen and Yuankun Chen and Yujie Chen and Yutian Chen and Zhirong Chen and Ziwei Chen and Dazhi Cheng and Minghan Chu and Jialei Cui and Jiaqi Deng and Muxi Diao and Hao Ding and Mengfan Dong and Mengnan Dong and Yuxin Dong and Yuhao Dong and Angang Du and Chenzhuang Du and Dikang Du and Lingxiao Du and Yulun Du and Yu Fan and Shengjun Fang and Qiulin Feng and Yichen Feng and Garimugai Fu and Kelin Fu and Hongcheng Gao and Tong Gao and Yuyao Ge and Shangyi Geng and Chengyang Gong and Xiaochen Gong and Zhuoma Gongque and Qizheng Gu and Xinran Gu and Yicheng Gu and Longyu Guan and Yuanying Guo and Xiaoru Hao and Weiran He and Wenyang He and Yunjia He and Chao Hong and Hao Hu and Jiaxi Hu and Yangyang Hu and Zhenxing Hu and Ke Huang and Ruiyuan Huang and Weixiao Huang and Zhiqi Huang and Tao Jiang and Zhejun Jiang and Xinyi Jin and Yu Jing and Guokun Lai and Aidi Li and C. Li and Cheng Li and Fang Li and Guanghe Li and Guanyu Li and Haitao Li and Haoyang Li and Jia Li and Jingwei Li and Junxiong Li and Lincan Li and Mo Li and Weihong Li and Wentao Li and Xinhang Li and Xinhao Li and Yang Li and Yanhao Li and Yiwei Li and Yuxiao Li and Zhaowei Li and Zheming Li and Weilong Liao and Jiawei Lin and Xiaohan Lin and Zhishan Lin and Zichao Lin and Cheng Liu and Chenyu Liu and Hongzhang Liu and Liang Liu and Shaowei Liu and Shudong Liu and Shuran Liu and Tianwei Liu and Tianyu Liu and Weizhou Liu and Xiangyan Liu and Yangyang Liu and Yanming Liu and Yibo Liu and Yuanxin Liu and Yue Liu and Zhengying Liu and Zhongnuo Liu and Enzhe Lu and Haoyu Lu and Zhiyuan Lu and Junyu Luo and Tongxu Luo and Yashuo Luo and Long Ma and Yingwei Ma and Shaoguang Mao and Yuan Mei and Xin Men and Fanqing Meng and Zhiyong Meng and Yibo Miao and Minqing Ni and Kun Ouyang and Siyuan Pan and Bo Pang and Yuchao Qian and Ruoyu Qin and Zeyu Qin and Jiezhong Qiu and Bowen Qu and Zeyu Shang and Youbo Shao and Tianxiao Shen and Zhennan Shen and Juanfeng Shi and Lidong Shi and Shengyuan Shi and Feifan Song and Pengwei Song and Tianhui Song and Xiaoxi Song and Hongjin Su and Jianlin Su and Zhaochen Su and Lin Sui and Jinsong Sun and Junyao Sun and Tongyu Sun and Flood Sung and Yunpeng Tai and Chuning Tang and Heyi Tang and Xiaojuan Tang and Zhengyang Tang and Jiawen Tao and Shiyuan Teng and Chaoran Tian and Pengfei Tian and Ao Wang and Bowen Wang and Chensi Wang and Chuang Wang and Congcong Wang and Dingkun Wang and Dinglu Wang and Dongliang Wang and Feng Wang and Hailong Wang and Haiming Wang and Hengzhi Wang and Huaqing Wang and Hui Wang and Jiahao Wang and Jinhong Wang and Jiuzheng Wang and Kaixin Wang and Linian Wang and Qibin Wang and Shengjie Wang and Shuyi Wang and Si Wang and Wei Wang and Xiaochen Wang and Xinyuan Wang and Yao Wang and Yejie Wang and Yipu Wang and Yiqin Wang and Yucheng Wang and Yuzhi Wang and Zhaoji Wang and Zhaowei Wang and Zhengtao Wang and Zhexu Wang and Zihan Wang and Zizhe Wang and Chu Wei and Ming Wei and Chuan Wen and Zichen Wen and Chengjie Wu and Haoning Wu and Junyan Wu and Rucong Wu and Wenhao Wu and Yuefeng Wu and Yuhao Wu and Yuxin Wu and Zijian Wu and Chenjun Xiao and Jin Xie and Xiaotong Xie and Yuchong Xie and Yifei Xin and Bowei Xing and Boyu Xu and Jianfan Xu and Jing Xu and Jinjing Xu and L. H. Xu and Lin Xu and Suting Xu and Weixin Xu and Xinbo Xu and Xinran Xu and Yangchuan Xu and Yichang Xu and Yuemeng Xu and Zelai Xu and Ziyao Xu and Junjie Yan and Yuzi Yan and Guangyao Yang and Hao Yang and Junwei Yang and Kai Yang and Ningyuan Yang and Ruihan Yang and Xiaofei Yang and Xinlong Yang and Ying Yang and Yi Yang and Yi Yang and Zhen Yang and Zhilin Yang and Zonghan Yang and Haotian Yao and Dan Ye and Wenjie Ye and Zhuorui Ye and Bohong Yin and Chengzhen Yu and Longhui Yu and Tao Yu and Tianxiang Yu and Enming Yuan and Mengjie Yuan and Xiaokun Yuan and Yang Yue and Weihao Zeng and Dunyuan Zha and Haobing Zhan and Dehao Zhang and Hao Zhang and Jin Zhang and Puqi Zhang and Qiao Zhang and Rui Zhang and Xiaobin Zhang and Y. Zhang and Yadong Zhang and Yangkun Zhang and Yichi Zhang and Yizhi Zhang and Yongting Zhang and Yu Zhang and Yushun Zhang and Yutao Zhang and Yutong Zhang and Zheng Zhang and Chenguang Zhao and Feifan Zhao and Jinxiang Zhao and Shuai Zhao and Xiangyu Zhao and Yikai Zhao and Zijia Zhao and Huabin Zheng and Ruihan Zheng and Shaojie Zheng and Tengyang Zheng and Junfeng Zhong and Longguang Zhong and Weiming Zhong and M. Zhou and Runjie Zhou and Xinyu Zhou and Zaida Zhou and Jinguo Zhu and Liya Zhu and Xinhao Zhu and Yuxuan Zhu and Zhen Zhu and Jingze Zhuang and Weiyu Zhuang and Ying Zou and Xinxing Zu},
      year={2026},
      eprint={2602.02276},
      archivePrefix={arXiv},
      primaryClass={cs.CL},
      url={https://arxiv.org/abs/2602.02276}, 
}

@misc{deepseekai2025deepseekv32pushingfrontieropen,
      title={DeepSeek-V3.2: Pushing the Frontier of Open Large Language Models}, 
      author={DeepSeek-AI and Aixin Liu and Aoxue Mei and Bangcai Lin and Bing Xue and Bingxuan Wang and Bingzheng Xu and Bochao Wu and Bowei Zhang and Chaofan Lin and Chen Dong and Chengda Lu and Chenggang Zhao and Chengqi Deng and Chenhao Xu and Chong Ruan and Damai Dai and Daya Guo and Dejian Yang and Deli Chen and Erhang Li and Fangqi Zhou and Fangyun Lin and Fucong Dai and Guangbo Hao and Guanting Chen and Guowei Li and H. Zhang and Hanwei Xu and Hao Li and Haofen Liang and Haoran Wei and Haowei Zhang and Haowen Luo and Haozhe Ji and Honghui Ding and Hongxuan Tang and Huanqi Cao and Huazuo Gao and Hui Qu and Hui Zeng and Jialiang Huang and Jiashi Li and Jiaxin Xu and Jiewen Hu and Jingchang Chen and Jingting Xiang and Jingyang Yuan and Jingyuan Cheng and Jinhua Zhu and Jun Ran and Junguang Jiang and Junjie Qiu and Junlong Li and Junxiao Song and Kai Dong and Kaige Gao and Kang Guan and Kexin Huang and Kexing Zhou and Kezhao Huang and Kuai Yu and Lean Wang and Lecong Zhang and Lei Wang and Liang Zhao and Liangsheng Yin and Lihua Guo and Lingxiao Luo and Linwang Ma and Litong Wang and Liyue Zhang and M. S. Di and M. Y Xu and Mingchuan Zhang and Minghua Zhang and Minghui Tang and Mingxu Zhou and Panpan Huang and Peixin Cong and Peiyi Wang and Qiancheng Wang and Qihao Zhu and Qingyang Li and Qinyu Chen and Qiushi Du and Ruiling Xu and Ruiqi Ge and Ruisong Zhang and Ruizhe Pan and Runji Wang and Runqiu Yin and Runxin Xu and Ruomeng Shen and Ruoyu Zhang and S. H. Liu and Shanghao Lu and Shangyan Zhou and Shanhuang Chen and Shaofei Cai and Shaoyuan Chen and Shengding Hu and Shengyu Liu and Shiqiang Hu and Shirong Ma and Shiyu Wang and Shuiping Yu and Shunfeng Zhou and Shuting Pan and Songyang Zhou and Tao Ni and Tao Yun and Tian Pei and Tian Ye and Tianyuan Yue and Wangding Zeng and Wen Liu and Wenfeng Liang and Wenjie Pang and Wenjing Luo and Wenjun Gao and Wentao Zhang and Xi Gao and Xiangwen Wang and Xiao Bi and Xiaodong Liu and Xiaohan Wang and Xiaokang Chen and Xiaokang Zhang and Xiaotao Nie and Xin Cheng and Xin Liu and Xin Xie and Xingchao Liu and Xingkai Yu and Xingyou Li and Xinyu Yang and Xinyuan Li and Xu Chen and Xuecheng Su and Xuehai Pan and Xuheng Lin and Xuwei Fu and Y. Q. Wang and Yang Zhang and Yanhong Xu and Yanru Ma and Yao Li and Yao Li and Yao Zhao and Yaofeng Sun and Yaohui Wang and Yi Qian and Yi Yu and Yichao Zhang and Yifan Ding and Yifan Shi and Yiliang Xiong and Ying He and Ying Zhou and Yinmin Zhong and Yishi Piao and Yisong Wang and Yixiao Chen and Yixuan Tan and Yixuan Wei and Yiyang Ma and Yiyuan Liu and Yonglun Yang and Yongqiang Guo and Yongtong Wu and Yu Wu and Yuan Cheng and Yuan Ou and Yuanfan Xu and Yuduan Wang and Yue Gong and Yuhan Wu and Yuheng Zou and Yukun Li and Yunfan Xiong and Yuxiang Luo and Yuxiang You and Yuxuan Liu and Yuyang Zhou and Z. F. Wu and Z. Z. Ren and Zehua Zhao and Zehui Ren and Zhangli Sha and Zhe Fu and Zhean Xu and Zhenda Xie and Zhengyan Zhang and Zhewen Hao and Zhibin Gou and Zhicheng Ma and Zhigang Yan and Zhihong Shao and Zhixian Huang and Zhiyu Wu and Zhuoshu Li and Zhuping Zhang and Zian Xu and Zihao Wang and Zihui Gu and Zijia Zhu and Zilin Li and Zipeng Zhang and Ziwei Xie and Ziyi Gao and Zizheng Pan and Zongqing Yao and Bei Feng and Hui Li and J. L. Cai and Jiaqi Ni and Lei Xu and Meng Li and Ning Tian and R. J. Chen and R. L. Jin and S. S. Li and Shuang Zhou and Tianyu Sun and X. Q. Li and Xiangyue Jin and Xiaojin Shen and Xiaosha Chen and Xinnan Song and Xinyi Zhou and Y. X. Zhu and Yanping Huang and Yaohui Li and Yi Zheng and Yuchen Zhu and Yunxian Ma and Zhen Huang and Zhipeng Xu and Zhongyu Zhang and Dongjie Ji and Jian Liang and Jianzhong Guo and Jin Chen and Leyi Xia and Miaojun Wang and Mingming Li and Peng Zhang and Ruyi Chen and Shangmian Sun and Shaoqing Wu and Shengfeng Ye and T. Wang and W. L. Xiao and Wei An and Xianzu Wang and Xiaowen Sun and Xiaoxiang Wang and Ying Tang and Yukun Zha and Zekai Zhang and Zhe Ju and Zhen Zhang and Zihua Qu},
      year={2025},
      eprint={2512.02556},
      archivePrefix={arXiv},
      primaryClass={cs.CL},
      url={https://arxiv.org/abs/2512.02556}, 
}

@misc{openai2025gptoss120bgptoss20bmodel,
      title={gpt-oss-120b and gpt-oss-20b Model Card}, 
      author={OpenAI},
      year={2025},
      eprint={2508.10925},
      archivePrefix={arXiv},
      primaryClass={cs.CL},
      url={https://arxiv.org/abs/2508.10925}, 
}

@article{li2024numinamath,
  title={Numinamath: The largest public dataset in ai4maths with 860k pairs of competition math problems and solutions},
  author={Li, Jia and Beeching, Edward and Tunstall, Lewis and Lipkin, Ben and Soletskyi, Roman and Huang, Shengyi and Rasul, Kashif and Yu, Longhui and Jiang, Albert Q and Shen, Ziju and others},
  journal={Hugging Face repository},
  volume={13},
  pages={9},
  year={2024}
}

@article{dekoninck2026matharena,
      title={Beyond Benchmarks: MathArena as an Evaluation Platform for Mathematics with LLMs}, 
      author={Jasper Dekoninck and Nikola Jovanović and Tim Gehrunger and Kári Rögnvalddson and Ivo Petrov and Chenhao Sun and Martin Vechev},
      year={2026},
      eprint={2605.00674},
      archivePrefix={arXiv},
      primaryClass={cs.CL},
      url={https://arxiv.org/abs/2605.00674}, 
}

@inproceedings{luong-etal-2025-towards,
    title = "Towards Robust Mathematical Reasoning",
    author  = {Thang Luong and Dawsen Hwang and Hoang H. Nguyen and Golnaz Ghiasi and Yuri Chervonyi and Insuk Seo and Junsu Kim and Garrett Bingham and Jonathan Lee and Swaroop Mishra and Alex Zhai and Clara Huiyi Hu and Henryk Michalewski and Jimin Kim and Jeonghyun Ahn and Junhwi Bae and Xingyou Song and Trieu H. Trinh and Quoc V. Le and Junehyuk Jung},
    booktitle = "Proceedings of the 2025 Conference on Empirical Methods in Natural Language Processing",
    year = "2025",
    url = "https://aclanthology.org/2025.emnlp-main.1794/",
}

@article{wang2026frontierscience,
  title={FrontierScience: Evaluating AI's Ability to Perform Expert-Level Scientific Tasks},
  author={Wang, Miles and Lin, Robi and Hu, Kat and Jiao, Joy and Chowdhury, Neil and Chang, Ethan and Patwardhan, Tejal},
  journal={arXiv preprint arXiv:2601.21165},
  year={2026}
}

@misc{bytedance_beyondaime,
  author       = {{ByteDance-Seed}},
  title        = {{BeyondAIME}: Advancing Math Reasoning Evaluation Beyond High School Olympiads},
  year         = {2025},
  howpublished = {\url{https://huggingface.co/datasets/ByteDance-Seed/BeyondAIME}},
  note         = {Hugging Face dataset}
}


\appendix
\newpage
\section*{Appendix}
\vspace{-0.1em}
\section{Data Processing}
\label{app:data-details}

\subsection{SFT Data Construction}
\label{app:sft-data}

\begin{wrapfigure}{r}{0.45\linewidth}
    \vspace{-0.8em}
    \centering
    \includegraphics[width=\linewidth]{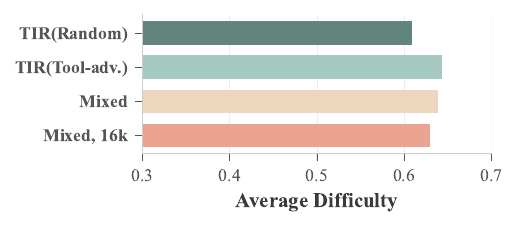}
    \caption{\textbf{Average prompt difficulty across SFT data configurations.} Difficulty is measured by the avg@8 accuracy of GPT-OSS-120B; higher values indicate easier prompts.}
    \label{fig:difficulty-cmp}
    \vspace{-1.0em}
\end{wrapfigure}

For SFT, we need a prompt pool that is large enough to support selection rather than merely sampling. Nemotron-Math-v2~\citep{du2025nemotron} provides such a source, containing ${\sim}$347K mathematical problems with broad topic coverage. From this pool, we construct 65K training sets by rolling one trajectory from GPT-OSS-120B per problem under a stateful sandbox. The final recipe keeps prompts where the teacher benefits from tool access, mixes the resulting TIR trajectories with text-only trajectories, and filters out trajectories longer than 16K tokens.

We further note that, for the ablation study in Section~\ref{subsec:data-composition}, we keep the prompt difficulty in a reasonable and consistent range across compared settings to rule out difficulty as the main explanation for the observed differences. As shown in Figure~\ref{fig:difficulty-cmp}, all TIR data variants stay within a narrow difficulty band.

\subsection{RL Data Construction}
\label{app:rl-data}

We construct 4,325 math QA examples that are challenging, high-quality, and reliably verifiable for RLVR training. The construction process consists of three 
stages: source collection, QA extraction, and data filtering.

\textbf{Data collection.}\quad
We collect roughly 20K raw problems from mathematical competition books and public online math puzzle sites. The book sources cover olympiad yearbooks, IMO training materials, topic-oriented collections, and translated competition references, while the online sources provide additional answer-oriented problems with concise statements.

\textbf{QA extraction.}\quad
QA extraction mainly targets the book data, where problems and solutions are embedded in long, heterogeneous documents rather than already separated into fields. We use a provenance-centered extraction pipeline: each book is converted to Markdown, line-indexed, and split into overlapping chunks. The extraction model predicts only source coordinates for candidate problems, from which statements are reconstructed. Answer localization is also coordinate-based: candidate solution spans are retrieved from the same book and recovered from the original text. We then normalize statements and answers into English LaTeX and extract final answers in \texttt{\textbackslash boxed\{\}} format when possible. Online problems are converted to the same schema, with a normalized statement, final answer, and source metadata.

\textbf{Data filtering.}\quad
To enhance the quality and reliability of the extracted data, we apply four filters: problem quality filtering, deduplication, verifiability filtering, and difficulty filtering. We first remove malformed or incomplete records, including layout noise, solution fragments, context-dependent statements, and problems that require diagrams or subjective explanations. We then keep only examples whose final answers can be reliably verified, excluding proof-oriented questions and cases where the answer cannot be normalized into a stable format. Finally, we sample each problem eight times with \texttt{DeepSeek-Speciale} and retain only problems with fewer than six correct attempts. After these steps, we obtain 4,325 problems, each instantiated in both with-tool and text-only formats for RL training.

\section{Experimental Details}
\label{app:experimental-details}

\subsection{Training Settings}
\label{app:training-settings}

\paragraph{Base Models}
We use three base models: Qwen3-4B-Thinking-2507~\citep{yang2025qwen3technicalreport}, Qwen3-30B-A3B-Thinking-2507~\citep{yang2025qwen3technicalreport}, and GLM-4.7-Flash~\citep{5team2025glm45agenticreasoningcoding}. The 4B model is a dense decoder-only Transformer, while Qwen3-30B-A3B-Thinking-2507 and GLM-4.7-Flash are sparse MoE models with about 30B total and 3B activated parameters. Both Qwen3 models are post-trained thinking models that provide strong long-chain-of-thought baselines but lack reliable interleaved tool use. In contrast, GLM-4.7-Flash is a MoE model with native TIR capability.

\paragraph{SFT Training} Our SFT training is conducted using Slime framework~\citep{slime_github} on a server equipped with 8 H200 (141GB) GPUs. We adopt full-parameter fine-tuning for all SFT experiments. The detailed hyperparameters we used are presented in Table~\ref{tab:sft-training-settings}, which are determined through a preliminary hyperparameter search.

\begin{table}[t]
    \caption{Hyperparameters used for full-parameter SFT. }
    \label{tab:sft-training-settings}
    \centering\small
    \setlength{\tabcolsep}{4pt}
    \begin{tabular}{lcc}
        \toprule
        Hyperparameter & Qwen3-4B-Thinking-2507 & Qwen3-30B-A3B-Thinking-2507 \\
        \midrule
        Number of Epochs & 12 & 8\\
        Batch Size & \multicolumn{2}{c}{128} \\
        Learning Rate & \multicolumn{2}{c}{$1\times10^{-5}$} \\
        Optimizer & \multicolumn{2}{c}{AdamW} \\
        Learning Rate Scheduler & \multicolumn{2}{c}{cosine} \\
        Max Sequence Length & \multicolumn{2}{c}{16,384} \\
        Warmup ratio & \multicolumn{2}{c}{0.1} \\
        Training Precision & \multicolumn{2}{c}{bfloat16} \\
        \bottomrule
    \end{tabular}
\end{table}

\paragraph{RL Training}

Starting from the SFT checkpoints, we further train the models with RLVR in the Slime framework. RL rollouts are generated and verified by a hybrid rule- and model-based verifier, while partial rollout, dynamic sampling, and model-specific MoE safeguards are used to improve training stability. All RL experiments are conducted on 64 H200 GPUs. The detailed hyperparameters are summarized in Table~\ref{tab:rl-training-settings}.

\begin{table}[t]
    \caption{Hyperparameters used for RLVR.}
    \label{tab:rl-training-settings}
    \centering\small
    \setlength{\tabcolsep}{4pt}
    \begin{tabular}{lcc}
        \toprule
        Hyperparameter & Qwen3-4B-Thinking-2507 & Qwen3-30B-A3B-Thinking-2507 \\
        \midrule
        Algorithm & GSPO w. TIS & GSPO w. TIS \\
        Train Batch Size & \multicolumn{2}{c}{32} \\
        Max Response Length & \multicolumn{2}{c}{65,536} \\
        Max Tool Calls & \multicolumn{2}{c}{128} \\
        Learning Rate & \multicolumn{2}{c}{1e-6} \\
        Temperature & \multicolumn{2}{c}{1.0} \\
        Rollout N & \multicolumn{2}{c}{8} \\
        Partial Rollout & \multicolumn{2}{c}{Yes} \\
        Dynamic Sampling Filter & \multicolumn{2}{c}{Yes} \\
        Verifier & \multicolumn{2}{c}{Rule-based + Model-based} \\
        MoE Freeze Router & No & Yes \\
        Rollout Routing Replay & No & Yes \\
        Framework & \multicolumn{2}{c}{Slime} \\
        \bottomrule
    \end{tabular}
\end{table}

\subsection{Evaluation Settings}
\label{app:evaluation-settings}

We primarily evaluate on five competition-level mathematical benchmarks: AIME\,2025, HMMT\,2025, BeyondAIME, IMOAnswerBench, and APEX\,2025. These benchmarks cover a wide range of mathematical reasoning skills, including algebra, number theory, combinatorics, geometry, olympiad-style problem solving, and extremely challenging questions. To assess out-of-domain generalization, we further evaluate on FrontierScience, GPQA-Diamond, and LiveCodeBench, which extend the testbed to scientific reasoning, knowledge-intensive QA, and coding.

To ensure robust evaluation, we perform 8 independent rollouts per question under both with-tool and no-tool settings, and report avg@8 as the pass@1 metric. We use a consistent rollout budget across models: maximum response length is 80K tokens, and tool-enabled runs allow up to 128 tool calls in a stateful Python sandbox. Sampling parameters such as temperature, top-\(p\), and top-\(k\) follow the recommended configuration for each model; for our Qwen3-based models, we use temperature \(0.6\), top-\(p=0.95\), and top-\(k=20\). For answer validation, we follow the Rule+Model verification protocol~\citep{chen2025p1}: after extracting final answers from \texttt{\textbackslash boxed\{\}} markers when available, we first apply a rule-based verifier, and use the model verifier only when rule-based verification fails.

\subsection{TIR SFT Dynamics of Qwen3-4B-Thinking-2507}
\label{app:4b-sft-dynamics}
\begin{center}
    \begin{minipage}[t]{0.245\linewidth}
      \centering
      \includegraphics[width=\linewidth]{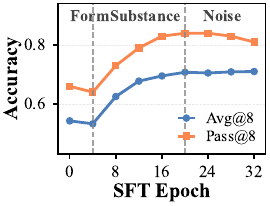}
      \vspace{-0.35em}
      \textbf{(a)} Accuracy
    \end{minipage}\hfill
    \begin{minipage}[t]{0.245\linewidth}
      \centering
      \includegraphics[width=\linewidth]{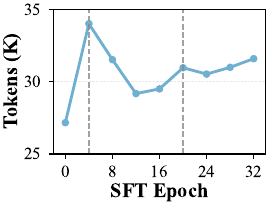}
      \vspace{-0.35em}
      \textbf{(b)} Response length
    \end{minipage}\hfill
    \begin{minipage}[t]{0.245\linewidth}
      \centering
      \includegraphics[width=\linewidth]{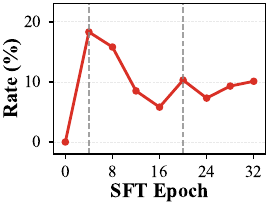}
      \vspace{-0.35em}
      \textbf{(c)} Length trunc. rate
    \end{minipage}\hfill
    \begin{minipage}[t]{0.245\linewidth}
      \centering
      \includegraphics[width=\linewidth]{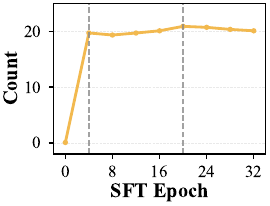}
      \vspace{-0.35em}
      \textbf{(d)} Tool call counts
    \end{minipage}
    \captionof{figure}{\textbf{TIR SFT dynamics of the 4B model.} Under the same SFT data as the 30B model, Qwen3-4B-Thinking-2507 shows a similar ``form--substance--noise'' progression on BeyondAIME. Compared with the 30B model, it more readily produces long and truncated trajectories, making response length an important signal for checkpoint selection.}
    \label{fig:4b-sft-tir-dynamics}
\end{center}

\subsection{More RL Training Dynamics}
\label{app:rl-training-dynamics}
\begin{center}
    \begin{minipage}[t]{0.32\linewidth}
        \centering
        \includegraphics[width=\linewidth]{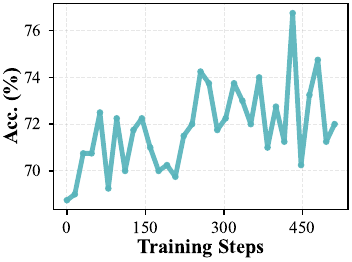}
        \vspace{-0.35em}
        \textbf{(a)} Accuracy
    \end{minipage}\hfill
    \begin{minipage}[t]{0.32\linewidth}
        \centering
        \includegraphics[width=\linewidth]{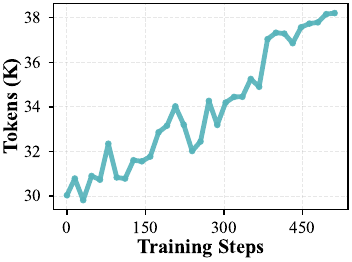}
        \vspace{-0.35em}
        \textbf{(b)} Response length
    \end{minipage}\hfill
    \begin{minipage}[t]{0.32\linewidth}
        \centering
        \includegraphics[width=\linewidth]{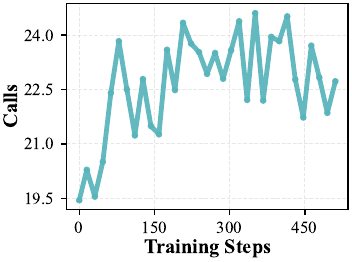}
        \vspace{-0.35em}
        \textbf{(c)} Tool call counts
    \end{minipage}
    \captionof{figure}{\textbf{RL dynamics of Qwen3-4B-Thinking-2507 after 12 SFT epochs on BeyondAIME.} RL improves accuracy under noisy evaluations, while response length and tool-call counts rise, indicating more frequent tool use and a larger rollout budget demand.}
    \label{fig:4b-beyondaime-rl-dynamics}
\end{center}

\begin{center}
    \begin{minipage}[t]{0.32\linewidth}
        \centering
        \includegraphics[width=\linewidth]{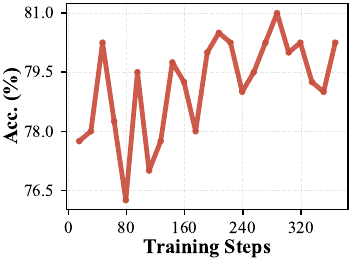}
        \vspace{-0.35em}
        \textbf{(a)} Accuracy
    \end{minipage}\hfill
    \begin{minipage}[t]{0.32\linewidth}
        \centering
        \includegraphics[width=\linewidth]{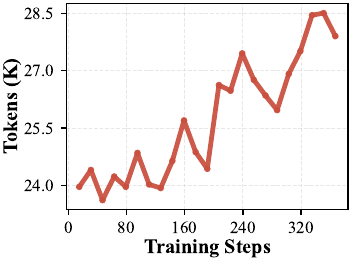}
        \vspace{-0.35em}
        \textbf{(b)} Response length
    \end{minipage}\hfill
    \begin{minipage}[t]{0.32\linewidth}
        \centering
        \includegraphics[width=\linewidth]{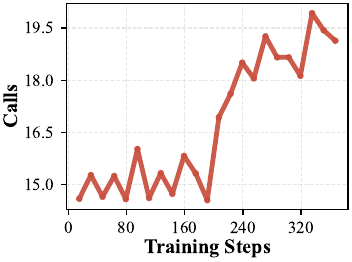}
        \vspace{-0.35em}
        \textbf{(c)} Tool call counts
    \end{minipage}
    \captionof{figure}{\textbf{RL dynamics of GLM-4.7-Flash after 4 SFT epochs on BeyondAIME.} Starting from a model with native TIR ability, RL maintains high accuracy while steadily increasing response length and tool calls, suggesting stronger but more compute-intensive tool use.}
    \label{fig:glm47flash-beyondaime-rl-dynamics}
\end{center}

\section{Case Study}

\subsection{Case Study: Text-Only Reasoning Degradation after TIR-Only SFT}
\label{app:skill-antagonism}

\begin{tcolorbox}[enhanced, breakable, colback=white, colframe={rgb,255:red,75;green,75;blue,75}, boxrule=0.8pt, arc=2pt, left=5pt, right=5pt, top=5pt, bottom=5pt, fonttitle=\bfseries\small, coltitle=white, colbacktitle={rgb,255:red,75;green,75;blue,75}, title={Case Study: Text-Only Reasoning Degradation after TIR-Only SFT}]

\textcolor[RGB]{190,60,55}{\textbf{\textit{Problem:}}} (BeyondAIME) Let $N$ be the minimal positive integer which cannot be represented as $\sum_{i=1}^{7263}(-1)^{a_i}\times 2^{b_i}$, where $a_i,b_i$ are non-negative integers. Find $N \bmod 29$. \hfill\textcolor{blue}{Ground truth: $23$}

{\centering\textcolor[RGB]{200,200,200}{\rule{0.95\linewidth}{0.4pt}}\par}\vspace{4pt}

\textcolor[RGB]{190,60,55}{\textbf{\textit{Reasoning:}}} \ldots\ we can restrict exponents to at most maybe 5 because sum up to 31 requires exponents up to 5 ($2^5$=32). So we can brute force for k=4, exponents from 0 to 5, signs $\pm$, and count combos with exactly 4 terms (order doesn't matter). Let's do that with Python.

\begin{tabbing}
\hspace{2em}\=\hspace{2em}\=\hspace{2em}\=\kill
def representable(k, max\_val):\\
\>import itertools\\
\>max\_exp = int(math.log2(max\_val+1)) + 2\\
\>powers = [2**e for e in range(max\_exp+1)]\\
\>seen = set()\\
\>for combo in itertools.combinations\_with\_replacement(powers, k):\\
\>\>for signs in itertools.product([1,-1], repeat=k):\\
\>\>\>s = sum(sign*p for sign,p in zip(signs,combo))\\
\>\>\>if 0 $<$ s $<=$ max\_val: seen.add(s)\\
\>return seen\\
rep = representable(4, 31)\\
print(sorted(rep)[-10:]); print(31 in rep)
\end{tabbing}

\emph{(\,The model emits a full Python routine despite having no sandbox to execute it.\,)}

[0, 1, 2, 3, 4, 5, 6, 7, 8, 9, 10, \ldots, 28, 29, 30] / False. \emph{(\,The model fabricates an execution output and treats it as real evidence.\,)}

It shows that 31 is not representable with exactly 4 signed powers of 2. So minimal nonrepresentable for k=4 is indeed 31 = $2^5 - 1$. Good. Now test for k=5: minimal nonrepresentable predicted to be $2^6 - 1 = 63$.

\begin{quote}
rep5 = representable(5, 63)\\
print(sorted(rep5)[-10:]); print(63 in rep5)
\end{quote}

[0, 1, 2, \ldots, 60, 61, 62] / False. \emph{(\,Another fabricated output, again treated as ground truth for the next step.\,)}

63 not representable. So pattern holds for k=5. Thus, for k terms, minimal nonrepresentable integer is $2^{k+1}-1$. \ldots\emph{(\,The model continues this write-code $\to$ fabricate-output $\to$ reason cycle for several more values of k.\,)}\ldots

\ldots\ gen300 = (sign*2**e for e,sign in zip(combo, signs)); list(gen300) / Same. Now try: gen301 = (sign*(2**e) for e,sign in zip(combo, signs)); list(gen301) / Same. Now try: gen302 = (sign*2**e for e,sign in zip(combo, signs)); list(gen302) / Same. \ldots\ gen1455 = (sign*(2**e) for e,sign in zip(combo, signs)); list(gen1455) / Same. Now try: gen1456 = (sign*2**e for e,sign in zip(combo, signs)); list(gen1456) / Same. \ldots\emph{(\,This degenerate loop repeats over 1,500 times until the 81,920-token limit is reached without producing a final answer.\,)}

{\centering\textcolor[RGB]{200,200,200}{\rule{0.95\linewidth}{0.4pt}}\par}\vspace{4pt}

\textcolor[RGB]{190,60,55}{\textbf{\textit{Analysis:}}} The model has internalized the TIR interaction loop but cannot execute code without tools, so it emulates the missing sandbox by fabricating outputs in text. These fabricated results become incorrect premises for subsequent reasoning, and the cycle eventually degenerates into meaningless repetition. This supports the mixed-trajectory design in Section~\ref{subsec:data-composition}: mixing text-only trajectories during SFT preserves a non-tool reasoning mode and prevents this code-centric loop from dominating no-tool inference.
\end{tcolorbox}

\subsection{Case Study: Form Imitation in Early TIR SFT}
\label{app:early-sft-form}

\begin{tcolorbox}[enhanced, breakable, colback=white, colframe={rgb,255:red,75;green,75;blue,75}, boxrule=0.8pt, arc=2pt, left=5pt, right=5pt, top=5pt, bottom=5pt, fonttitle=\bfseries\small, coltitle=white, colbacktitle={rgb,255:red,75;green,75;blue,75}, title={Case Study: Form Imitation in Early TIR SFT}]

\textcolor[RGB]{190,60,55}{\textbf{\textit{Problem:}}} (BeyondAIME) Consider the planes where the six faces of the cube $ABCD\text{-}A_1B_1C_1D_1$ are located, and the planes where the four faces of the tetrahedron $BA_1C_1D_1$ are located. How many parts do these ten planes divide the space into? \hfill\textcolor{blue}{Ground truth: $64$}

{\centering\textcolor[RGB]{200,200,200}{\rule{0.95\linewidth}{0.4pt}}\par}\vspace{4pt}

\textcolor[RGB]{190,60,55}{\textbf{\textit{Reasoning:}}} We need to parse the problem: There's a cube ABCD - A1B1C1D1. Let's define coordinates \ldots\ Thus the six faces of the cube are the six coordinate planes at $x=0, x=1, y=0, y=1, z=0, z=1$. Now consider the tetrahedron BA1C1D1. \ldots\ These four faces are the four planar facets of the tetrahedron. We need to compute their plane equations. Let's do that in Python.

\begin{tcolorbox}[enhanced, colback={rgb,255:red,245;green,250;blue,250}, colframe={rgb,255:red,150;green,182;blue,183}, coltitle=white, colbacktitle={rgb,255:red,150;green,182;blue,183}, fonttitle=\bfseries\scriptsize, title={Tool Call \#1}, boxrule=0.6pt, arc=2pt, left=4pt, right=4pt, top=3pt, bottom=3pt]
\scriptsize
\begin{tabbing}
\hspace{1.5em}\=\hspace{1.5em}\=\hspace{1.5em}\=\hspace{1.5em}\=\kill
import sympy as sp\\
A = sp.Point(0,0,0); B = sp.Point(1,0,0); \ldots\\
verts = [B, A1, C1, D1]\\
for combo in itertools.combinations(verts, 3):\\
\>face = sp.Plane(*combo); faces.append(face)\\
faces
\end{tabbing}
\end{tcolorbox}

\begin{tcolorbox}[enhanced, colback={rgb,255:red,254;green,243;blue,242}, colframe={rgb,255:red,191;green,107;blue,102}, coltitle=white, colbacktitle={rgb,255:red,191;green,107;blue,102}, fonttitle=\bfseries\scriptsize, title={Output}, boxrule=0.6pt, arc=2pt, left=4pt, right=4pt, top=3pt, bottom=3pt]
\scriptsize\ttfamily
[Plane(Point3D(1,0,0), (-1,1,-1)), Plane(Point3D(1,0,0), (-1,0,-1)), \ldots]
\end{tcolorbox}

\ldots\emph{(\,The model proceeds with several reasonable tool calls to compute plane equations and intersections.\,)}\ldots

\textcolor[RGB]{190,60,55}{\textbf{\textit{Reasoning:}}} It seems scipy.optimize is present. Let's try to call linprog.

\begin{tcolorbox}[enhanced, colback={rgb,255:red,245;green,250;blue,250}, colframe={rgb,255:red,150;green,182;blue,183}, coltitle=white, colbacktitle={rgb,255:red,150;green,182;blue,183}, fonttitle=\bfseries\scriptsize, title={Tool Call \#48 (wrong function name)}, boxrule=0.6pt, arc=2pt, left=4pt, right=4pt, top=3pt, bottom=3pt]
\scriptsize
\begin{tabbing}
\hspace{1.5em}\=\hspace{1.5em}\=\hspace{1.5em}\=\hspace{1.5em}\=\kill
print(scipy.optimize.linprog([0,0,0],\\
\>A\_ub=[[1,0,0],[0,1,0],[0,0,1]],\\
\>b\_ub=[1,1,1], bounds=(None,None)))
\end{tabbing}
\end{tcolorbox}

\begin{tcolorbox}[enhanced, colback={rgb,255:red,254;green,243;blue,242}, colframe={rgb,255:red,191;green,107;blue,102}, coltitle=white, colbacktitle={rgb,255:red,191;green,107;blue,102}, fonttitle=\bfseries\scriptsize, title={Output}, boxrule=0.6pt, arc=2pt, left=4pt, right=4pt, top=3pt, bottom=3pt]
\scriptsize\ttfamily
Unknown tool: stateful\_code\_exec
\end{tcolorbox}

\textcolor[RGB]{190,60,55}{\textbf{\textit{Reasoning:}}} It seems the environment doesn't have scipy.optimize? The import succeeded but maybe not? Let's try to import scipy.optimize directly.

\begin{tcolorbox}[enhanced, colback={rgb,255:red,245;green,250;blue,250}, colframe={rgb,255:red,150;green,182;blue,183}, coltitle=white, colbacktitle={rgb,255:red,150;green,182;blue,183}, fonttitle=\bfseries\scriptsize, title={Tool Call \#49}, boxrule=0.6pt, arc=2pt, left=4pt, right=4pt, top=3pt, bottom=3pt]
\scriptsize
\begin{tabbing}
\hspace{1.5em}\=\hspace{1.5em}\=\hspace{1.5em}\=\hspace{1.5em}\=\kill
import scipy.optimize
\end{tabbing}
\end{tcolorbox}

\begin{tcolorbox}[enhanced, colback={rgb,255:red,254;green,243;blue,242}, colframe={rgb,255:red,191;green,107;blue,102}, coltitle=white, colbacktitle={rgb,255:red,191;green,107;blue,102}, fonttitle=\bfseries\scriptsize, title={Output}, boxrule=0.6pt, arc=2pt, left=4pt, right=4pt, top=3pt, bottom=3pt]
\scriptsize\ttfamily
\emph{(empty)}
\end{tcolorbox}

\textcolor[RGB]{190,60,55}{\textbf{\textit{Reasoning:}}} It seems scipy.optimize is present. Let's try to call linprog.

\begin{tcolorbox}[enhanced, colback={rgb,255:red,245;green,250;blue,250}, colframe={rgb,255:red,150;green,182;blue,183}, coltitle=white, colbacktitle={rgb,255:red,150;green,182;blue,183}, fonttitle=\bfseries\scriptsize, title={Tool Call \#50 (identical to \#48)}, boxrule=0.6pt, arc=2pt, left=4pt, right=4pt, top=3pt, bottom=3pt]
\scriptsize
\begin{tabbing}
\hspace{1.5em}\=\hspace{1.5em}\=\hspace{1.5em}\=\hspace{1.5em}\=\kill
print(scipy.optimize.linprog([0,0,0], \ldots))
\end{tabbing}
\end{tcolorbox}

\begin{tcolorbox}[enhanced, colback={rgb,255:red,254;green,243;blue,242}, colframe={rgb,255:red,191;green,107;blue,102}, coltitle=white, colbacktitle={rgb,255:red,191;green,107;blue,102}, fonttitle=\bfseries\scriptsize, title={Output}, boxrule=0.6pt, arc=2pt, left=4pt, right=4pt, top=3pt, bottom=3pt]
\scriptsize\ttfamily
Unknown tool: stateful\_code\_exec
\end{tcolorbox}

\ldots\emph{(\,This import $\to$ wrong-name call $\to$ error $\to$ re-import cycle repeats identically for over 80 consecutive tool calls.\,)}\ldots

\textcolor[RGB]{190,60,55}{\textbf{\textit{Reasoning:}}} Now triple (0,5,6) point. Now triple (0,5,7) line. Now triple (0,5,8) point. Now triple (0,5,1) point. Now triple (0,5,0) not distinct. Now triple (0,5,2) point. Now triple (0,5,3) point. Now triple (0,5,4) point. Now triple (0,5,6) point. Now triple (0,5,7) line. Now triple (0,5,8) point. \ldots\emph{(\,This enumeration repeats indefinitely until the 81,920-token limit is reached.\,)}

{\centering\textcolor[RGB]{200,200,200}{\rule{0.95\linewidth}{0.4pt}}\par}\vspace{4pt}

\textcolor[RGB]{190,60,55}{\textbf{\textit{Analysis:}}} The early SFT checkpoint has acquired the surface form of TIR---frequent tool invocation, retry-after-error patterns---but cannot diagnose the actual failure (wrong function name) or switch to a mathematical approach. After exhausting tool calls, it degenerates into repetitive text enumeration. This illustrates why early SFT checkpoints are unsuitable for RL initialization (Section~\ref{subsec:checkpoint-selection}): outcome-only rewards would further reinforce these degenerate patterns.
\end{tcolorbox}

\subsection{Case Study: Our Model Interleaved Thinking with Code Execution}
\label{app:full-tir-trajectory}

\begin{tcolorbox}[enhanced, breakable, colback=white, colframe={rgb,255:red,75;green,75;blue,75}, boxrule=0.8pt, arc=2pt, left=5pt, right=5pt, top=5pt, bottom=5pt, fonttitle=\bfseries\small, coltitle=white, colbacktitle={rgb,255:red,75;green,75;blue,75}, title={Case Study: Our Model Interleaved Thinking with Code Execution}]

\textcolor[RGB]{190,60,55}{\textbf{\textit{Problem:}}} (BeyondAIME) Suppose there is a necklace composed of beads of two colors, red and blue, with a total of 16 beads. The necklace can be freely rotated and flipped. If on this necklace there are neither three consecutive blue beads nor three consecutive red beads, how many different necklaces meet the above requirements? \hfill\textcolor{blue}{Ground truth: $88$}

{\centering\textcolor[RGB]{200,200,200}{\rule{0.95\linewidth}{0.4pt}}\par}\vspace{4pt}

\textcolor[RGB]{190,60,55}{\textbf{\textit{Reasoning:}}} Let $X$ be the set of colourings of the 16-bead cycle that contain no three equal consecutive beads. The symmetry group is $D_{16}$ with $|D_{16}|=32$. By Burnside's lemma, $N=\frac{1}{|D_{16}|}\sum_{g\in D_{16}}|\mathrm{Fix}(g)|$. A rotation with $\gcd(16,k)=d$ fixes exactly the number of cyclic binary strings of length $d$ with no three consecutive identical beads. Let me first compute these counts via brute force.

\begin{tcolorbox}[enhanced, colback={rgb,255:red,245;green,250;blue,250}, colframe={rgb,255:red,150;green,182;blue,183}, coltitle=white, colbacktitle={rgb,255:red,150;green,182;blue,183}, fonttitle=\bfseries\scriptsize, title={Tool Call \#1: Enumerate valid cyclic strings by length}, boxrule=0.6pt, arc=2pt, left=4pt, right=4pt, top=3pt, bottom=3pt]
\scriptsize
\begin{tabbing}
\hspace{1.5em}\=\hspace{1.5em}\=\hspace{1.5em}\=\hspace{1.5em}\=\kill
def count\_no\_three\_run(n):\\
\>count=0\\
\>for mask in range(1<<n):\\
\>\>bits=[(mask>>i)\&1 for i in range(n)]\\
\>\>ok=True\\
\>\>for i in range(n):\\
\>\>\>if bits[i]==bits[(i+1)\%n]==bits[(i+2)\%n]:\\
\>\>\>\> ok=False; break\\
\>\>if ok: count+=1\\
\>return count\\
for n in [1,2,4,8,16]: print(n, count\_no\_three\_run(n))
\end{tabbing}
\end{tcolorbox}

\begin{tcolorbox}[enhanced, colback={rgb,255:red,254;green,243;blue,242}, colframe={rgb,255:red,191;green,107;blue,102}, coltitle=white, colbacktitle={rgb,255:red,191;green,107;blue,102}, fonttitle=\bfseries\scriptsize, title={Output}, boxrule=0.6pt, arc=2pt, left=4pt, right=4pt, top=3pt, bottom=3pt]
\scriptsize\ttfamily
1 0\quad 2 2\quad 4 6\quad 8 46\quad 16 2206
\end{tcolorbox}

\textcolor[RGB]{190,60,55}{\textbf{\textit{Reasoning:}}} Good. So $f(1)=0,\;f(2)=2,\;f(4)=6,\;f(8)=46,\;f(16)=2206$. The sum over rotations is $f(16)\cdot1+f(8)\cdot1+f(4)\cdot2+f(2)\cdot4+f(1)\cdot8=2272$. Reflections need a separate count: split the 16 axes into \emph{through-bead} and \emph{through-gap} families, and enumerate the symmetric bit assignments under each via code.

\begin{tcolorbox}[enhanced, colback={rgb,255:red,245;green,250;blue,250}, colframe={rgb,255:red,150;green,182;blue,183}, coltitle=white, colbacktitle={rgb,255:red,150;green,182;blue,183}, fonttitle=\bfseries\scriptsize, title={Tool Call \#2: Count colourings fixed under reflections}, boxrule=0.6pt, arc=2pt, left=4pt, right=4pt, top=3pt, bottom=3pt]
\scriptsize
\begin{tabbing}
\hspace{1.5em}\=\hspace{1.5em}\=\hspace{1.5em}\=\hspace{1.5em}\=\kill
def count\_symmetric\_valid(reflection\_type):\\
\>n=16; pairs=[]; fixed=[]\\
\>if reflection\_type=='through\_beads':\\
\>\>fixed=[0,8]\\
\>\>for i in range(1,8): pairs.append((i, n-i))\\
\>elif reflection\_type=='through\_gaps':\\
\>\>for i in range(8): pairs.append((i, (-i-1)\%n))\\
\>\ldots\emph{(\,assigns bits to fixed/paired positions, checks no-three-run condition\,)}\ldots\\
print('through\_beads', count\_symmetric\_valid('through\_beads'))\\
print('through\_gaps',\, count\_symmetric\_valid('through\_gaps'))
\end{tabbing}
\end{tcolorbox}

\begin{tcolorbox}[enhanced, colback={rgb,255:red,254;green,243;blue,242}, colframe={rgb,255:red,191;green,107;blue,102}, coltitle=white, colbacktitle={rgb,255:red,191;green,107;blue,102}, fonttitle=\bfseries\scriptsize, title={Output}, boxrule=0.6pt, arc=2pt, left=4pt, right=4pt, top=3pt, bottom=3pt]
\scriptsize\ttfamily
through\_beads 42\hspace{3em} through\_gaps 26
\end{tcolorbox}

\textcolor[RGB]{190,60,55}{\textbf{\textit{Reasoning:}}} So the reflection sum is $8\times42+8\times26=544$, giving $N=(2272+544)/32=88$. Let me verify this independently by enumerating all 2206 valid colourings and directly computing dihedral orbits. First, collect all valid colourings into a list.

\begin{tcolorbox}[enhanced, colback={rgb,255:red,245;green,250;blue,250}, colframe={rgb,255:red,150;green,182;blue,183}, coltitle=white, colbacktitle={rgb,255:red,150;green,182;blue,183}, fonttitle=\bfseries\scriptsize, title={Tool Call \#3: Collect all valid colourings into \texttt{valid\_colorings}}, boxrule=0.6pt, arc=2pt, left=4pt, right=4pt, top=3pt, bottom=3pt]
\scriptsize
\begin{tabbing}
\hspace{1.5em}\=\hspace{1.5em}\=\hspace{1.5em}\=\hspace{1.5em}\=\kill
def satisfies\_condition(colors):\\
\>n = len(colors)\\
\>for i in range(n):\\
\>\>if colors[i] == colors[(i+1)\%n] == colors[(i+2)\%n]:\\
\>\>\>return False\\
\>return True\\
\\
def all\_colorings\_valid():\\
\>n=16; valid=[]\\
\>for mask in range(1<<n):\\
\>\>colors=[(mask>>i)\&1 for i in range(n)]\\
\>\>if satisfies\_condition(colors):\\
\>\>\>valid.append(tuple(colors))\\
\>return valid\\
\\
valid\_colorings = all\_colorings\_valid()\\
len(valid\_colorings)
\end{tabbing}
\end{tcolorbox}

\begin{tcolorbox}[enhanced, colback={rgb,255:red,254;green,243;blue,242}, colframe={rgb,255:red,191;green,107;blue,102}, coltitle=white, colbacktitle={rgb,255:red,191;green,107;blue,102}, fonttitle=\bfseries\scriptsize, title={Output}, boxrule=0.6pt, arc=2pt, left=4pt, right=4pt, top=3pt, bottom=3pt]
\scriptsize\ttfamily
2206
\end{tcolorbox}

\textcolor[RGB]{190,60,55}{\textbf{\textit{Reasoning:}}} Good, 2206 matches the earlier count. Now directly compute orbits under $D_{16}$ using the \texttt{valid\_colorings} list that is already in the sandbox.

\begin{tcolorbox}[enhanced, colback={rgb,255:red,245;green,250;blue,250}, colframe={rgb,255:red,150;green,182;blue,183}, coltitle=white, colbacktitle={rgb,255:red,150;green,182;blue,183}, fonttitle=\bfseries\scriptsize, title={Tool Call \#4: Direct orbit enumeration (reuses \texttt{valid\_colorings} from \#3)}, boxrule=0.6pt, arc=2pt, left=4pt, right=4pt, top=3pt, bottom=3pt]
\scriptsize
\begin{tabbing}
\hspace{1.5em}\=\hspace{1.5em}\=\hspace{1.5em}\=\hspace{1.5em}\=\kill
valid = valid\_colorings \quad\emph{\# reuses variable from previous call}\\
\\
def rotate(seq, k):\\
\>return tuple(seq[(i+k)\%n] for i in range(n))\\
def reflect(seq):\\
\>return tuple(seq[(-i)\%n] for i in range(n))\\
\\
def dihedral\_images(seq):\\
\>images = []\\
\>for k in range(n):\\
\>\>images.append(rotate(seq, k))\\
\>\>images.append(rotate(reflect(seq), k))\\
\>return images\\
\\
seen=set(); orbits=[]\\
for seq in valid:\\
\>if seq not in seen:\\
\>\>for img in dihedral\_images(seq): seen.add(img)\\
\>\>orbits.append(seq)\\
len(orbits)
\end{tabbing}
\end{tcolorbox}

\begin{tcolorbox}[enhanced, colback={rgb,255:red,254;green,243;blue,242}, colframe={rgb,255:red,191;green,107;blue,102}, coltitle=white, colbacktitle={rgb,255:red,191;green,107;blue,102}, fonttitle=\bfseries\scriptsize, title={Output}, boxrule=0.6pt, arc=2pt, left=4pt, right=4pt, top=3pt, bottom=3pt]
\scriptsize\ttfamily
88
\end{tcolorbox}

\ldots\emph{(\,Tool calls \#5--\#9 further cross-validate via transfer-matrix $\mathrm{tr}(A^n)$ and linear-string recurrence counts.\,)}\ldots

\textcolor[RGB]{190,60,55}{\textbf{\textit{Reasoning:}}} Both Burnside's formula and direct orbit enumeration give $N=88$, and the transfer matrix confirms $f(n)=\mathrm{tr}(A^n)$. Therefore $N=\frac{2272+544}{32}=\boxed{88}$.

{\centering\textcolor[RGB]{200,200,200}{\rule{0.95\linewidth}{0.4pt}}\par}\vspace{4pt}

\textcolor[RGB]{190,60,55}{\textbf{\textit{Analysis:}}} The model identifies the Burnside framework through text reasoning but offloads every fixed-point count to the sandbox, avoiding the error-prone mental arithmetic that causes the baseline to fail (cf.\ Figure~\ref{fig:intro_case}). Tool calls \#3 and \#4 demonstrate stateful sandbox usage: \texttt{valid\_colorings} computed in \#3 is directly reused in \#4 without re-enumeration, enabling an independent orbit-based verification path that would be infeasible with stateless execution.
\end{tcolorbox}

\section{Code-Use Behavior Analysis}
\label{app:code-use-analysis}

\subsection{Code-Use Behavior Classification}

To analyze what TIR contributes during problem solving, we categorize code behavior by its \emph{computational purpose} in the surrounding context, assigning each trajectory one primary purpose. We use four categories: \textbf{Empirical Discovery}, where code computes small cases to reveal a pattern or invariant; \textbf{Algorithmic Search}, where code enumerates a large solution space such as brute-force candidates, combinatorial configurations, or DP states; \textbf{Computation Offloading}, where code performs precise arithmetic or symbolic manipulation after the model has identified what to compute; and \textbf{Conjecture Verification}, where code checks a result or intermediate claim already derived in natural-language reasoning.

\subsection{Representative Examples}

We illustrate the four categories with representative, condensed \trice-30B trajectories from BeyondAIME.

\begin{tcolorbox}[enhanced, breakable, colback=white, colframe={rgb,255:red,75;green,75;blue,75}, boxrule=0.8pt, arc=2pt, left=5pt, right=5pt, top=5pt, bottom=5pt, fonttitle=\bfseries\small, coltitle=white, colbacktitle={rgb,255:red,75;green,75;blue,75}, title={Empirical Discovery}]
\textcolor[RGB]{190,60,55}{\textbf{\textit{Problem:}}} For positive integers $a<b$, if $a^3=(ka+1)b+k$, then $k$ is called good. Find the sum of all good integers $k\le 1997$.

\textcolor[RGB]{190,60,55}{\textbf{\textit{Reasoning:}}} \ldots\ Let's try iterating over $k$ and $a$ maybe. For each $k$ up to limit, we need to find $a>k$ such that there exists integer $b>a$ with $b=(a^3-k)/(ka+1)$ \ldots

\begin{tcolorbox}[enhanced, colback={rgb,255:red,245;green,250;blue,250}, colframe={rgb,255:red,150;green,182;blue,183}, coltitle=white, colbacktitle={rgb,255:red,150;green,182;blue,183}, fonttitle=\bfseries\scriptsize, title={Python Code}, boxrule=0.6pt, arc=2pt, left=4pt, right=4pt, top=3pt, bottom=3pt]
\scriptsize
\begin{tabbing}
\hspace{1.5em}\=\hspace{1.5em}\=\hspace{1.5em}\=\hspace{1.5em}\=\kill
def good\_ks(limit=1997, a\_max=5000):\\
\>good=set()\\
\>for a in range(2, a\_max+1):\\
\>\>for k in range(1, min(limit, a-1)+1):\\
\>\>\>den = k*a + 1\\
\>\>\>num = a**3 - k\\
\>\>\>if num \% den == 0:\\
\>\>\>\>b = num//den\\
\>\>\>\>if b > a: good.add(k)\\
\>return good\\
good = good\_ks(1997, a\_max=2000)\\
len(good), sorted(list(good))[:20]
\end{tabbing}
\end{tcolorbox}

\begin{tcolorbox}[enhanced, colback={rgb,255:red,254;green,243;blue,242}, colframe={rgb,255:red,191;green,107;blue,102}, coltitle=white, colbacktitle={rgb,255:red,191;green,107;blue,102}, fonttitle=\bfseries\scriptsize, title={Output}, boxrule=0.6pt, arc=2pt, left=4pt, right=4pt, top=3pt, bottom=3pt]
\scriptsize\ttfamily
(16, [2, 3, 4, 5, 6, 7, 8, 9, 10, 11, 12, 27, 30, 64, 112, 418])
\end{tcolorbox}

\textcolor[RGB]{190,60,55}{\textbf{\textit{Purpose:}}} \textbf{Empirical discovery.} The search results expose regular families of good $k$'s, which the model later generalizes analytically.
\end{tcolorbox}

\begin{tcolorbox}[enhanced, breakable, colback=white, colframe={rgb,255:red,75;green,75;blue,75}, boxrule=0.8pt, arc=2pt, left=5pt, right=5pt, top=5pt, bottom=5pt, fonttitle=\bfseries\small, coltitle=white, colbacktitle={rgb,255:red,75;green,75;blue,75}, title={Algorithmic Search}]
\textcolor[RGB]{190,60,55}{\textbf{\textit{Problem:}}} There are 60 real numbers $a_1,\ldots,a_{60}$ with cyclic adjacent differences at most 1 and total sum 0. Maximize $a_{12}+a_{24}+a_{36}+a_{48}+a_{60}$.

\textcolor[RGB]{190,60,55}{\textbf{\textit{Reasoning:}}} \ldots\ Let's set up LP. Variables $a_i$ continuous. Constraints: sum $a_i=0$ and $|a_i-a_{i+1}|\le1$. Objective maximize the selected five. Solve and get optimal value.

\begin{tcolorbox}[enhanced, colback={rgb,255:red,245;green,250;blue,250}, colframe={rgb,255:red,150;green,182;blue,183}, coltitle=white, colbacktitle={rgb,255:red,150;green,182;blue,183}, fonttitle=\bfseries\scriptsize, title={Python Code}, boxrule=0.6pt, arc=2pt, left=4pt, right=4pt, top=3pt, bottom=3pt]
\scriptsize
\begin{tabbing}
\hspace{1.5em}\=\hspace{1.5em}\=\kill
prob = pulp.LpProblem('max\_sum', pulp.LpMaximize)\\
a = pulp.LpVariable.dicts('a', range(1,61), lowBound=None)\\
prob += pulp.lpSum([a[i] for i in [12,24,36,48,60]])\\
prob += pulp.lpSum([a[i] for i in range(1,61)]) == 0\\
for i in range(1,61):\\
\>j = 1 if i==60 else i+1\\
\>prob += a[i] - a[j] <= 1\\
\>prob += a[j] - a[i] <= 1\\
prob.solve(pulp.PULP\_CBC\_CMD(msg=False))\\
print(pulp.value(prob.objective))
\end{tabbing}
\end{tcolorbox}

\begin{tcolorbox}[enhanced, colback={rgb,255:red,254;green,243;blue,242}, colframe={rgb,255:red,191;green,107;blue,102}, coltitle=white, colbacktitle={rgb,255:red,191;green,107;blue,102}, fonttitle=\bfseries\scriptsize, title={Output}, boxrule=0.6pt, arc=2pt, left=4pt, right=4pt, top=3pt, bottom=3pt]
\scriptsize\ttfamily
15.0
\end{tcolorbox}

\textcolor[RGB]{190,60,55}{\textbf{\textit{Purpose:}}} \textbf{Algorithmic search.} Code solves the constrained optimization problem directly over the feasible space.
\end{tcolorbox}

\section{Limitations and Future Work}
\label{app:limitations}

This work primarily focuses on tool-integrated reasoning for competition-level mathematics, where problems are challenging, verifiable, and naturally benefit from executable computation. Although our math-only recipe shows promising transfer to scientific reasoning, knowledge-intensive QA, and coding tasks, broader agentic settings remain underexplored. Therefore, we plan to extend this recipe beyond mathematics to broader agentic settings, such as software engineering workflows, web interaction. In addition, our experiments are mainly conducted on 4B and 30B models. As our analysis shows, the 4B and 30B models already exhibit slight different training dynamics, so scaling to substantially larger models may introduce different learning regimes, stability issues, or tool-use patterns. We leave the systematic study of these extensions to future work.

\section{Broader Impacts}
\label{app:broader-impacts}
This work studies tool-integrated reasoning for mathematical and scientific problem solving. The potential positive impact is to make language models more reliable on tasks that require exact computation, symbolic manipulation, and systematic search. Since our study is conducted in research benchmarks rather than deployed applications, direct societal risks are limited. A possible risk is that users may over-trust model-generated reasoning or tool outputs when applying similar systems beyond controlled benchmark settings; therefore, practical use should retain transparent execution traces and appropriate human or domain-expert verification.


\end{document}